\documentclass[10pt,twocolumn,letterpaper]{article}

\usepackage[pagenumbers]{cvpr} 

\usepackage{siunitx}
\usepackage{tikz, pgfplots}

\definecolor{cvprblue}{rgb}{0.21,0.49,0.74}
\usepackage[pagebackref,breaklinks,colorlinks,citecolor=cvprblue]{hyperref}

\def\paperID{***} 
\def\confName{3DV\xspace}
\def\confYear{2024\xspace}

\title{Randomized 3D Scene Generation for Generalizable Self-Supervised Pre-Training}

\author{Lanxiao Li \quad\quad Michael Heizmann \\
Karlsruhe Institute of Technology, Karlsruhe, Germany \\
{\tt\small lanxiao.li@kit.edu} \quad\quad {\tt\small michael.heizmann@kit.edu}}

\begin{document}
\maketitle

\begin{abstract}
Capturing and labeling real-world 3D data is laborious and time-consuming, which makes it costly to train strong 3D models. To address this issue, recent works present a simple method by generating randomized 3D scenes without simulation and rendering. Although models pre-trained on the generated synthetic data gain impressive performance boosts, previous works have two major shortcomings. First, they focus on only one downstream task (i.e., object detection), and the generalization to other tasks is unexplored. Second, the contributions of generated data are not systematically studied. To obtain a deeper understanding of the randomized 3D scene generation technique, we revisit previous works and compare different data generation methods using a unified setup. Moreover, to clarify the generalization of the pre-trained models, we evaluate their performance in multiple tasks (i.e., object detection and semantic segmentation) and with different pre-training methods (i.e., masked autoencoder and contrastive learning). Moreover, we propose a new method to generate 3D scenes with spherical harmonics. It surpasses the previous formula-driven method with a clear margin and achieves on-par results with methods using real-world scans and CAD models. 
\end{abstract}

\section{Introduction}
\label{sec:intro}
Deep neural networks are data-hungry while capturing and labeling data requires significant time and human effort. This problem is especially concerning in 3D computer vision, as 3D data and labels are more scarce and expensive. 
To train strong 3D neural networks at a lower cost, many works apply synthetic data in pre-training and fine-tune the models on real-world data. 
One possible approach for generating synthetic data is simulation~\cite{deschaud:2021:paris_carla, griffiths:2019:synthcity, hurl:2019:presil_dataset, wu:2018:squeezeseg}. Although realistic scenes can be simulated, developing the simulation environment, crafting the source materials, and designing scene layouts still require much effort. 

\begin{figure}[ht]
	\centering
	\includegraphics[width=0.9\linewidth]{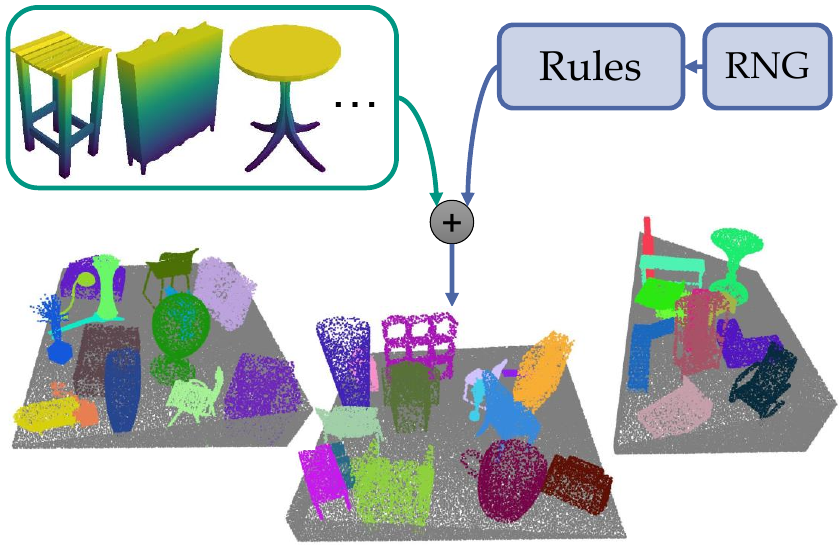}
	\caption{Concept of randomized 3D scene generation. Objects from an object set are randomly placed based on pre-defined rules. RNG: random number generator. }
	\label{fig:datagen_teaser}
\end{figure}

Recently, randomized 3D scene generation~\cite{rao:2021:randomrooms, yamada:2022:fractal_point} has shown promising results. This approach generates 3D scenes by randomly placing ``objects'', which can be CAD models~\cite{rao:2021:randomrooms} or formula-driven shapes~\cite{yamada:2022:fractal_point}, based on pre-defined rules. The concept of this approach is visualized in Figure~\ref{fig:datagen_teaser}. 
Randomized 3D scene generation requires neither real-world data nor manual annotation. Also, it is simpler and more efficient compared to simulation methods. 

However, previous works using this technique~\cite{rao:2021:randomrooms, yamada:2022:fractal_point} have two major issues. 
Firstly, their pre-training methods are task-specific. They are only designed and evaluated in the case of object detection, which limits their application to other tasks. Also, it is unclear if the learned features can be transferred to other tasks. 
Secondly, previous works apply novel data-generating methods and unique pre-training methods simultaneously. Therefore, the performance gains from the pre-training are brought by two factors, and the synthetic data's contributions are unclear. 
One may question whether the elaborately designed pre-training methods are more important than the synthetic data. 
Also, whether the generated data can work with other general~(and more widely applied) pre-training methods is unknown. 

To better understand the impact of the generated data, this work rethinks the randomized 3D scene generation technique by separating the influences of pre-training methods and applied data. 
Moreover, we study the generalization of features learned from the generated data. 
Specifically, we test the synthetic data with different pre-training methods, \ie, mask autoencoder and contrastive learning. 
Furthermore, we evaluate the pre-trained models for multiple downstream tasks, \ie, 3D object detection and semantic segmentation. 

Our experimental results demonstrate the power of randomized 3D scene generation: the generated data apply to different pre-training methods, and the pre-trained models gain significant improvement in different downstream tasks. In some tasks, \eg, 3D object detection on the ScanNet dataset using a VoteNet, the synthetic pre-training data even outperform real-world ones with the same data amount. 

On the other hand, using formula-driven shapes, \eg, fractal point clouds~\cite{yamada:2022:fractal_point}, instead of CAD models for scene generation is more promising since it does not rely on additional data and saves the cost of crafting or gathering CAD models. 
However, this work demonstrates that fractal point clouds lead to sub-optimal results in pre-training. We hypothesize that it is caused by a domain gap since the appearance of fractal point clouds is not close to real-world objects~(see Figure~\ref{fig:objects}). 
Instead of fractal point clouds, we propose using spherical harmonics as objects, which are formula-driven with natural surfaces. 
Their appearance is closer to real-world objects compared to fractal point clouds. 
Experimental results show that spherical harmonics outperform fractal point clouds with a clear margin and are competitive compared to CAD models and real-world data in pre-training.

The contribution of this work is many-fold: 
\begin{enumerate}
    \item We systematically study the randomized 3D scene generation technique and provide a deeper understanding of the generated data, \eg, the impact of object sets and view angles.  
    
    \item We study two aspects of generalization of the feature learning using the generated data, \ie, the generalization to different pre-training methods and multiple downstream tasks. 
    
    \item We propose a new method that generates scenes with formula-driven spherical harmonics. The models pre-trained using this approach achieve competitive results in downstream tasks without using real-world data or CAD models in the pre-training stage. 
\end{enumerate}

\section{Related Works}
This section explains the self-supervised pre-training for 3D computer vision and the simulation-based methods for generating synthetic data. Randomized 3D scene generation is discussed in detail in Section~\ref{subsec:revisit}. 

\noindent
\textbf{Pre-Training in 3D Vision.} A lot of methods utilize the invariance of 3D features to establish correspondence between different views and apply contrastive learning~\cite{xie:2020:pointcontrast, zhang:2021:depthcontrast, hou:2021:exploring_efficient, xu:2022:image2point, li:2022:closer}. Also, some works transfer knowledge from other domains, \eg, color images~\cite{qian:2022:pix4point} and language~\cite{zhang:2022:pointclip}. Moreover, some works~\cite{yu:2022:point_bert, fu:2022:pos_bert, wang:2021:occlusion} reconstruct partially visual point clouds in the pretext task. 
Recent works~\cite{pang:2022:point_mae, zhang:2022:point_m2ae, liu:2022:masked_discrimi, li:2023:applying} apply the successful masked autoencoder~\cite{he:2022:mae} on point clouds. 
However, these works pre-train models using either hand-crafted CAD models~\cite{wu:2015:modelnet_dataset, chang:2015:shapenet_dataset} or real-world point scans~\cite{dai:2017:scannet_dataset}. Their application to generated synthetic data is under-explored. 

\noindent
\textbf{Simulation-Based Scene Generation.} 
Deschaud~\etal~\cite{deschaud:2021:paris_carla} employ CARLA simulator~\cite{Dosovitskiy:17:carla_simulator} and generate synthetic LiDAR point clouds for 3D mapping tasks. 
Griffiths and Boehm~\cite{griffiths:2019:synthcity} present a large-scale point cloud of the urban environment using BlenSor~\cite{gschwandtner:2011:blensor}. 
Xiao~\etal~\cite{xiao:2022:transfer_from_synthetic} collect point cloud data using Unreal Engine~4 (UE4) and investigate the transfer learning from synthetic to real-world data in semantic segmentation. 
Wu~\etal~\cite{wu:2018:squeezeseg} and Hurl~\etal~\cite{hurl:2019:presil_dataset} synthesize training data using Grand Theft Auto V, a popular video game. 

\section{Method}
In this section, we first briefly revisit the concept of randomized 3D scene generation and previous works. Then, we present details on our data generation process, including a novel method to create formula-driven shapes using spherical harmonics. 
Finally, we explain the self-supervised pre-training with generated data. 

\subsection{Revisiting Previous Works}
\label{subsec:revisit}
As shown in Figure~\ref{fig:datagen_teaser}, randomized 3D scene generation requires an object set and pre-defined rules. A room with a random size is first created to generate a new scene. A 3D object is randomly picked from the set, undergoes data augmentation (\eg, rotation and scaling), and is randomly placed in the room. This process is repeated until the room contains enough objects. 
The rules define \eg, the distribution of the room size, choices and parameters for data augmentation, distribution of objects in the room, and the number of objects in each scene. With an object set and rules given, a vast amount of scenes can be easily generated. 
Then, the generated data can be applied to pre-train 3D neural networks. 

\begin{figure}[hbt]
    \centering
    \setlength\tabcolsep{2pt}
    \begin{tabular}{cccc}
         \includegraphics[width=0.2\linewidth]{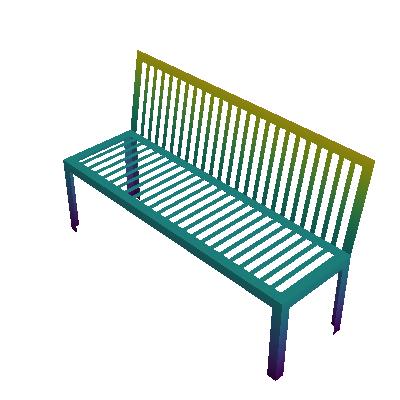} &
         \includegraphics[width=0.13\linewidth]{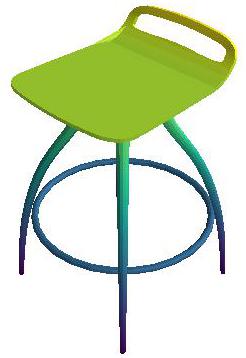} &
         \includegraphics[width=0.2\linewidth]{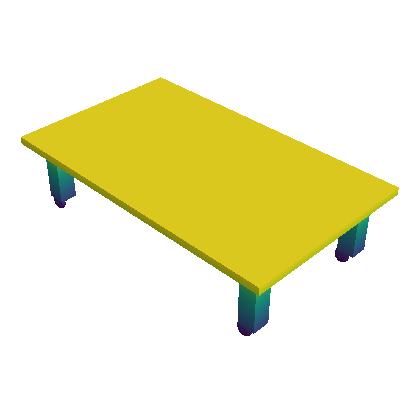} &
         \includegraphics[width=0.15\linewidth]{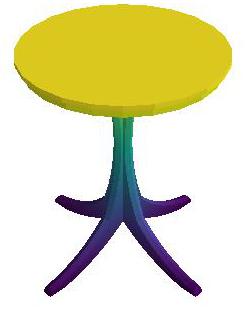} \\
         \includegraphics[width=0.2\linewidth]{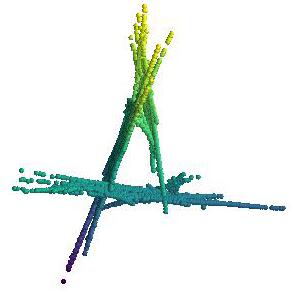} &
         \includegraphics[width=0.2\linewidth]{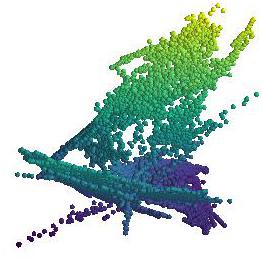} &
         \includegraphics[width=0.17\linewidth]{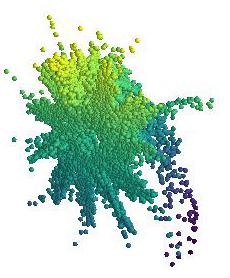} &
         \includegraphics[width=0.2\linewidth]{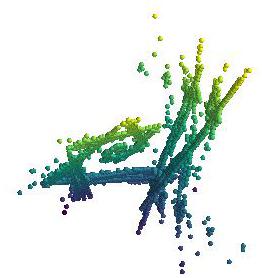} \\
         \includegraphics[width=0.2\linewidth]{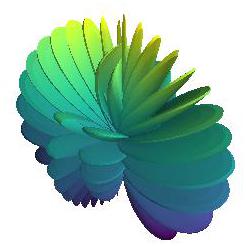} &
         \includegraphics[width=0.2\linewidth]{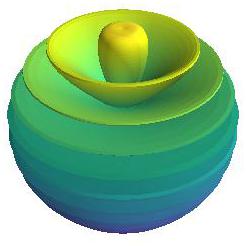} &
         \includegraphics[width=0.14\linewidth]{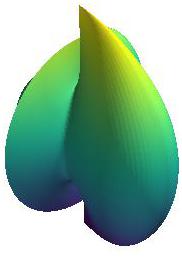} &
         \includegraphics[width=0.2\linewidth]{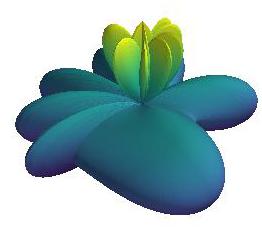} \\
    \end{tabular}
    \caption{Examples of objects for scene generation. First row: CAD models from ModelNet40~\cite{wu:2015:modelnet_dataset}. Second row: fractal point clouds from PC-FractalDB~\cite{yamada:2022:fractal_point}. Third row: spherical harmonics. CAD models and spherical harmonics are shown as meshes, although only sampled points are applied in pre-training. The color indicates the relative height. }
    \label{fig:objects}
\end{figure}

Previous works on randomized 3D scene generation use similar rules while having different choices on object sets. Rao~\etal~\cite{rao:2021:randomrooms} use CAD models, which is straightforward since openly accessible datasets of CAD models are commonly used in 3D computer vision~\cite{wu:2015:modelnet_dataset, chang:2015:shapenet_dataset}. This choice has the drawback that creating or gathering CAD models is still laborious and time-consuming. 
On the contrary, Yamada~\etal~\cite{yamada:2022:fractal_point} propose generating the object set randomly. Specifically, they create fractal point clouds by randomly sampling affine transformations and iteratively applying them to 3D points. 
However, the appearance of the generated point clouds is not close to real-world objects. 

Moreover, previous works apply unique pre-training methods, respectively. 
Rao~\etal~\cite{rao:2021:randomrooms} create pairs of scenes with the same objects but different room layouts and perform contrastive learning using the object-level correspondence. Yamada~\etal~\cite{yamada:2022:fractal_point} generate bounding box labels for each object and pre-train object detectors in a fully supervised manner. For the object classification sub-task, fractal point clouds are automatically labeled according to their spatial variance. 
Both methods are only evaluated in object detection. 

Although these two works have shown promising results, many questions remain open, \eg, do the randomly generated data apply to other downstream tasks? What is the contribution of the generated data in the pre-training? Do synthetic data still work without the elaborately designed pre-training methods? How does the choice of object sets affect the performance? 
This work takes a closer look at randomized 3D scene generation and attempts to answer these questions. 

\subsection{Spherical Harmonics as Objects}
As mentioned in Section~\ref{sec:intro}, using formula-driven fractal point clouds is more efficient than hand-crafted CAD models. However, their appearance is not similar to real-world objects. 
Moreover, since real-world 3D data are usually captured with depth sensors (\eg, laser scanners or RGB-D cameras), which cannot measure the internal structure of objects, real-world data only contain sample points on object surfaces. 
It implies that the objects used for scene generation should contain sufficient surfaces so that the pre-trained models can be transferred to downstream tasks with real-world data. 
Fractal point clouds, on the contrary, do not contain continuous surfaces (see Figure~\ref{fig:objects}). 
We hypothesize that the lack of surfaces also makes the pre-training less effective. 

We propose generating object sets using spherical harmonics, which can be represented in a spherical coordinate system with radial distance $r$, polar angle $\theta$, and azimuthal angle $\phi$: 
\begin{align}
    r =& \sin (m_1 \phi )^{p_1} + \cos (m_2 \phi )^{p_2} \nonumber \\
    & + \sin (m_3 \theta )^{p_3} + \cos (m_4 \theta )^{p_4} \, ,
\label{eq:harmonics}
\end{align}
where $m_i \in \mathbb{R}$ and $p_i \in \mathbb{Z^*}$ with $i \in \left \{1, 2, 3, 4 \right \}$. 
With fixed coefficients $m_i$ and $p_i$, Equation~\ref{eq:harmonics} describes a closed surface in 3D space. 
Diverse 3D objects can be generated when the coefficients are randomly set. Some generated spherical harmonics are visualized in Figure~\ref{fig:objects}. 
Spherical harmonics are initially employed to solve partial differential equations~\cite{muller:2006:spherical}. In computer vision, they are also applied to describe surfaces and shapes~\cite{kazhdan:2003:rotation, saupe:2001:3d}. The form in Equation~\ref{eq:harmonics} is motivated by the original definition but does not strictly follow it. 
This work does not consider the mathematical and physical meanings of spherical harmonics. They are only used to parameterize the object set. 
This idea is also inspired by a web page written by Paul Bourke\footnote{\url{http://paulbourke.net/geometry/sphericalh/}. Last accessed on 2023.04.03.}. 
The generated spherical harmonics can be easily represented as (rectangular) meshes when coordinates $\phi$ and $\theta$ are sampled with constant intervals. 
The mesh representation greatly simplifies further processing, \eg, the point sampling and ray-casting (explained in Section~\ref{subsec:single_view}).

\noindent
\textbf{Discussion on other choices.}
There are more choices to generate objects, \eg, using other parametric shapes~\cite{hobolth:2003:continuous, monedero:2000:parametric, schulz:2017:retrieval, smirnov:2020:deep} or generative deep learning models~\cite{nichol:2022:pointe, jun:2023:shap}. 
We use spherical harmonics because we believe their simplicity can highlight the effectiveness of randomized 3D scene generation. 
Furthermore, the simplicity could inspire more research interest of feature works: if spherical harmonics, described by only one simple equation, can achieve good performance, applying more sophisticated parametric shapes and their mixtures should be promising. 

\subsection{From Objects to Scenes}

This work assumes that all scenes are static. 
Also, only downstream tasks in indoor scenes are considered, following previous works~\cite{rao:2021:randomrooms, yamada:2022:fractal_point}. However, extending the methods to outdoor scenarios is straightforward. 
We adopt the generation rules of Rao~\etal~\cite{rao:2021:randomrooms} and represent generated 3D scenes as point clouds. 
Each randomly picked object is normalized into a unit sphere. 
Then, 3000 points are randomly sampled and undergo random data augmentation. Each generated scene contains 12 to 16 objects. 
Since random scaling is used as data augmentation, the point density on each object might differ. To make the point density consistent across each scene, grid sampling (\ie, voxelization) with the voxel size of \SI{0.04}{\meter} is applied to the point clouds. 

\begin{figure}[t]
	\centering
	\setlength\tabcolsep{1pt}
	\begin{tabular}{cccc}
		\includegraphics[width=0.30\linewidth]{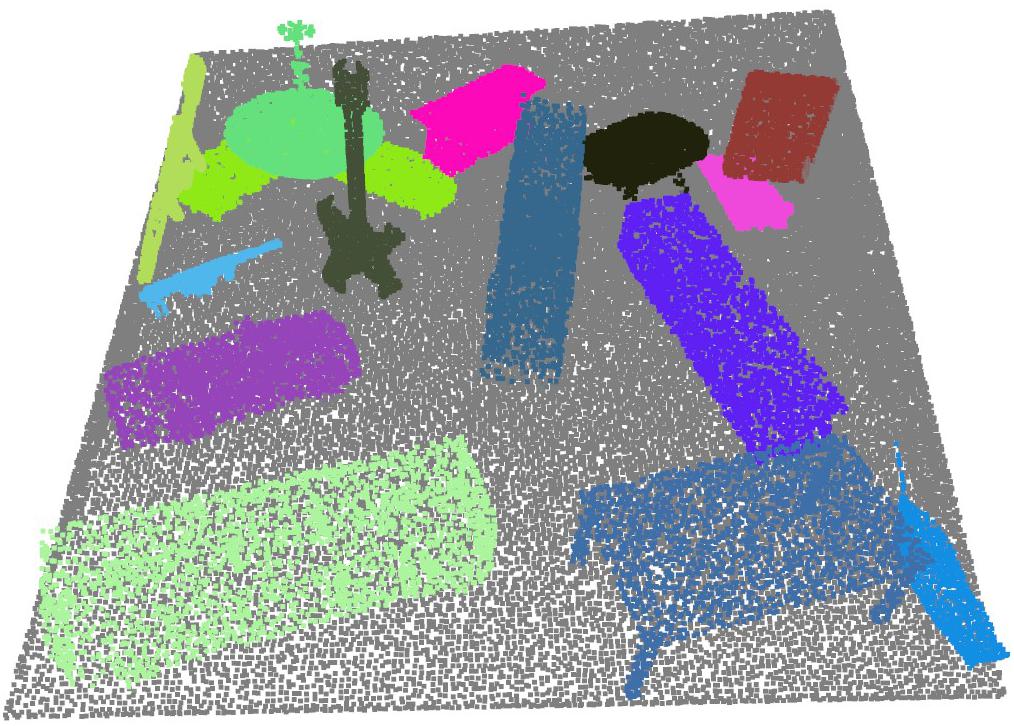} &
		\includegraphics[width=0.32\linewidth]{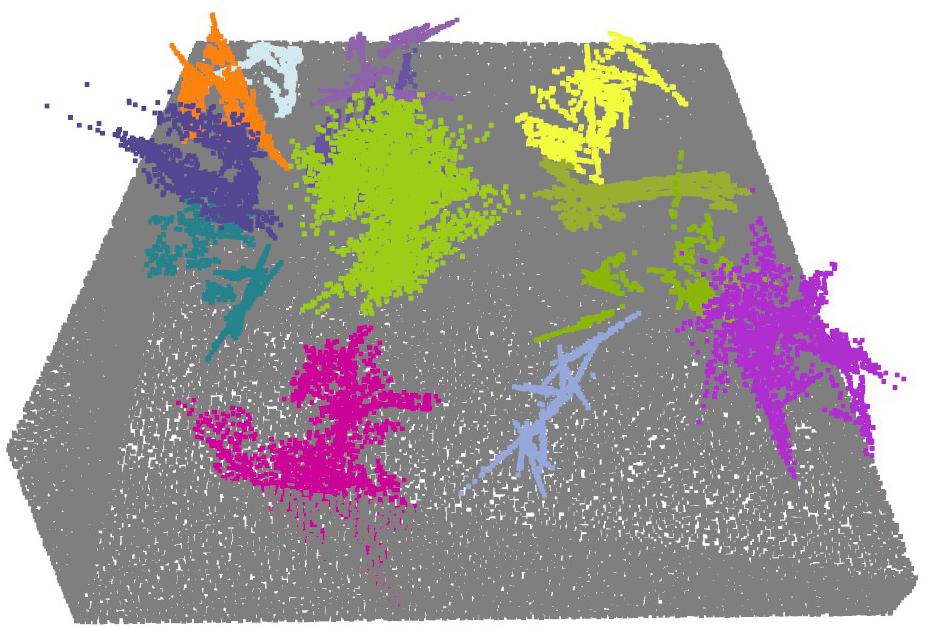} & 
		\includegraphics[width=0.32\linewidth]{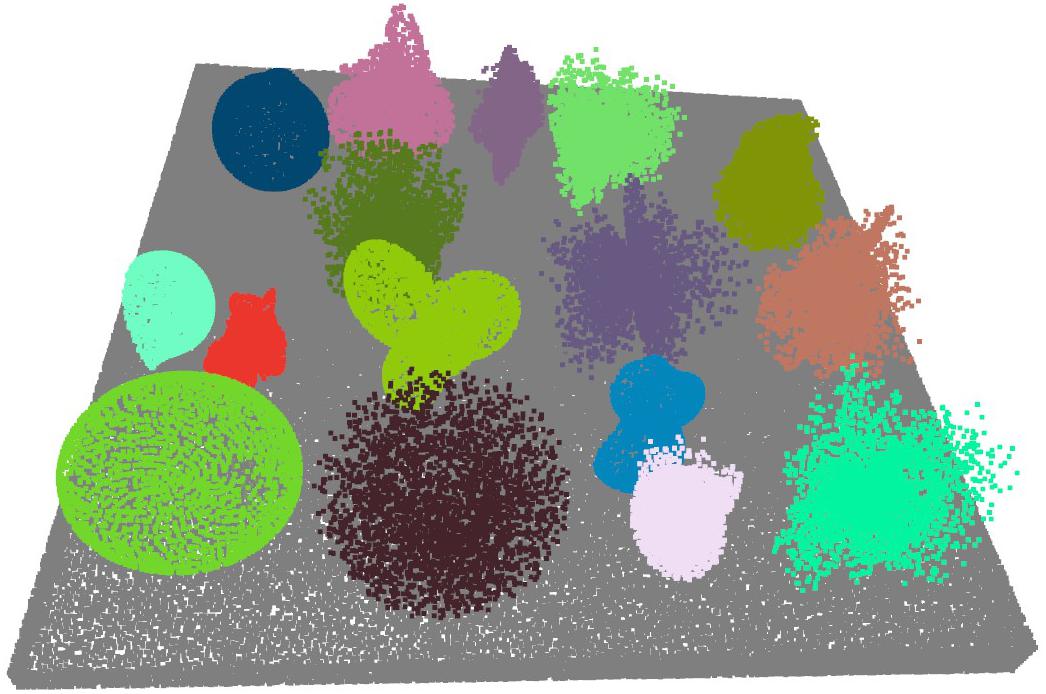}
		\\ (a) & (b) & (c)
	\end{tabular}
	\caption{Examples of generated 3D scenes in point cloud representation. (a): using CAD models from ShapeNet~\cite{chang:2015:shapenet_dataset}. (b): using fractal point cloud. (c): using spherical harmonics. The colors here are solely for visualization purposes. The generated synthetic point clouds do not contain color channels. Scenes using CAD models from ModelNet40~\cite{wu:2015:modelnet_dataset} are not illustrated as they are visually similar to the variants using ShapeNet.}
	\label{fig:scenes_pcd}
\end{figure}

The same rules are employed for all object sets (\eg, CAD models and fractal points). 
Some generated scenes are visualized in Figure~\ref{fig:scenes_pcd}. More details on generation rules and more visualizations are provided in our supplementary material. 
Note that we only generate single point clouds independently. 
On the contrary, Rao \etal~\cite{rao:2021:randomrooms} create pairs of point clouds with object-level correspondence, and Yamada~\etal~\cite{yamada:2022:fractal_point} additionally generate bounding box labels for each scene. 

\subsection{Single-View Point Clouds}
\label{subsec:single_view}

Previous works represent the generated scenes as multi-view point clouds. In this work, a point cloud is referred to as multi-view when it cannot be projected to one surface without information loss. 
A typical example of multi-view point clouds is a 3D scan reconstructed from a data sequence, \eg, using the SLAM (simultaneous location and mapping) technology. 
On the contrary, many point clouds in practice are single-view and represented as depth maps or range images. 
Besides comparing different methods for randomized scene generation, it is also meaningful to compare the synthetic data with real-world ones. Many works use single-view real-world data for self-supervised pre-training~\cite{zhang:2021:depthcontrast, xie:2020:pointcontrast, li:2022:closer, hou:2021:exploring_efficient, xu:2022:image2point} since they are the direct output from depth sensors and easy to obtain. 

\begin{figure}[hbt]
	\centering
	\setlength\tabcolsep{1pt}
	\begin{tabular}{ccc}
		\includegraphics[width=0.32\linewidth]{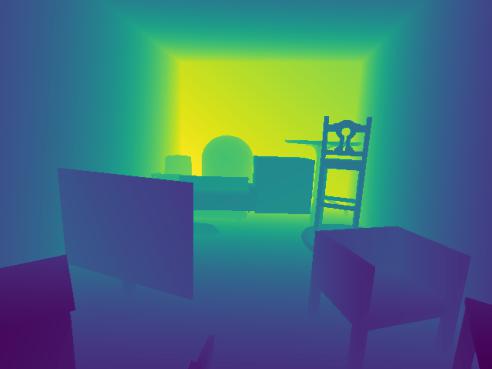} & 
		\includegraphics[width=0.32\linewidth]{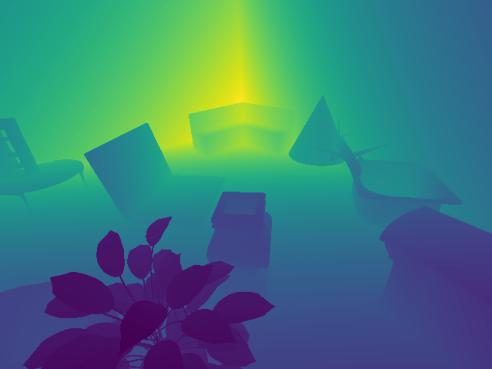} &
		\includegraphics[width=0.32\linewidth]{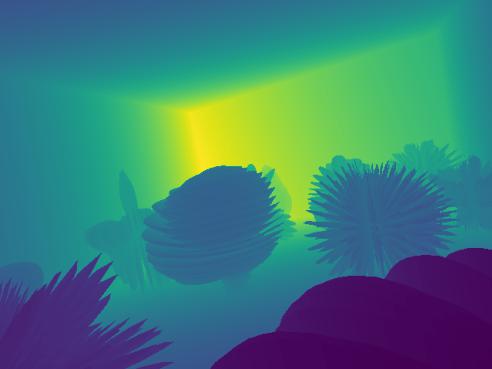}
	\\ (a) & (b) & (c)
	\end{tabular}
	\caption{Examples of generated depth maps via ray-casting. (a): with objects from ShapeNet~\cite{chang:2015:shapenet_dataset}. (b): with objects from ModelNet40~\cite{wu:2015:modelnet_dataset}. (c): with spherical harmonics. According to the rules in~\cite{rao:2021:randomrooms}, objects might be stacked on others. Depth values are normalized for visualization. }
	\label{fig:depth}
\end{figure}

To clarify the impact of single-view and multi-view representations in pre-training and for a fairer comparison with the real-world data, this work also studies single-view point clouds. 
Specifically, meshes are used as objects to create scenes instead of sampled points.
The same rules as in~\cite{rao:2021:randomrooms} are used to generate a 3D scene. 
Then, a virtual camera with a random pose captures a depth map using ray-casting~\cite{roth:1982:raycast}. 
The camera pose is checked so that each depth map contains sufficient objects (empirically set to 7 in this work). 
In pre-training, the depth maps are converted into point clouds on the flight. 
Data augmentations, \eg, cropping the depth maps, and scaling and rotating the point clouds converted from them, are also applied. 
Some generated depth maps are visualized in Figure~\ref{fig:depth}. No depth maps are generated with fractal point clouds since ray-casting is non-trivial in this case.

\subsection{Self-Supervised Pre-Training}

This work explores generalizable pre-training using generated synthetic data. Instead of using a pre-training strategy specific to one downstream task, we apply masked autoencoder (MAE)~\cite{he:2022:mae} and MoCoV2~\cite{chen:2020:mocov2} to evaluate the generated data, since they generalize well in multiple downstream tasks and have shown good results with real-world data~\cite{li:2023:applying, liu:2022:masked_discrimi, zhang:2021:depthcontrast, li:2022:closer}. 
The concepts of the two pre-training methods are briefly revisited in the following.

MAE~\cite{he:2022:mae} is employed to pre-train vision transformers~\cite{dosovitskiy:2020:vit, vaswani:2017:transformer}. 
An input point cloud is divided into small patches, and a large proportion (\eg, \SI{75}{\percent}) is masked. A transformer-based encoder extracts patch-wise features from the remaining patches, and a decoder is trained to reconstruct the masked patches. After the pre-training, the decoder is abandoned, and the encoder can be transferred into downstream tasks. 
We adopt the architecture and training method in~\cite{li:2023:applying} for MAE. 

Also, scene-level contrastive learning is applied to pre-train a point cloud encoder following~\cite{zhang:2021:depthcontrast}. Two overlapping regions are cropped from each generated scene to create a positive pair (\ie, an anchor and a positive sample). 
Point clouds from other scenes are regarded as negative samples. 
Given a point cloud as an anchor, the encoder is trained to distinguish its positive sample from many negative samples. This work applies MoCoV2~\cite{chen:2020:mocov2} by using an exponentially moving averaged momentum encoder and a memory bank to save global features from negative samples. 

\section{Experiments}
This section first introduces the experimental setups in this work. Then, it demonstrates the main results by comparing different synthetic datasets. Furthermore, the proposed methods are compared with state-of-the-art methods using synthetic or real-world data for pre-training. Finally, additional experimental results are provided to justify the design choices and demonstrate the effectiveness of the proposed methods. 

\subsection{Setups}
\label{subsec:setup}
Our default setups are explained in the following. 

\noindent
\textbf{Real-World Baseline.}
To create a baseline using real-world data, the models are pre-trained on ScanNet~\cite{dai:2017:scannet_dataset}, a large-scale indoor dataset captured in approximately 1500 rooms. 78K frames are sampled (approximately one in every 25 frames) from the raw data in the training set, following setups in previous works on pre-training~\cite{xie:2020:pointcontrast, hou:2021:exploring_efficient, li:2023:applying, li:2022:closer}. 

\noindent
\textbf{Scene Generation.}
To evaluate the impact of different object sets, four object sets are used to generate synthetic data, summarized in Table~\ref{tab:datagen_data}. 
ShapeNet~\cite{chang:2015:shapenet_dataset} and ModelNet40~\cite{wu:2015:modelnet_dataset} are used as CAD model sets. The former consists of $\sim$50K models of ordinary everyday objects (\eg, chairs and shelves), while the latter contains $\sim$10K. This work follows their official train/test split and uses only objects from training sets. 
In the default setup, 10K formula-driven objects are used to generate scenes, so the object number is aligned with ModelNet40. 
The configuration in~\cite{yamada:2022:fractal_point} is used to create fractal point clouds. 
For spherical harmonics, the coefficients $m_i$ and $p_i$ in Equation~\ref{eq:harmonics} are constrained in the range of $[-5,~5]$ and $[0,~4]$, respectively. 

\begin{table}[b]
	\centering
	\begin{tabular}{lcc}
		\toprule
		\textbf{Dataset} & \textbf{Obj. Set} & \textbf{Obj. Number} \\
		\hline
		RM-ShapeNet & ShapeNet~\cite{chang:2015:shapenet_dataset} & $\sim$50K\\
		RM-ModelNet & ModelNet40~\cite{wu:2015:modelnet_dataset} & $\sim$10K\\
		RM-Fractal & fractal points~\cite{yamada:2022:fractal_point} & 10K\\
		RM-Harmonics & spherical harmonics & 10K \\
		\bottomrule
	\end{tabular}
	\caption{Summary of generated datasets.}
	\label{tab:datagen_data}
\end{table}

By default, 78K scenes are generated with each object set, following the real-world baseline. The generated datasets are named RM-ShapeNet, RM-ModelNet, RM-Fractal, and RM-Harmonics according to object sets. 
The abbreviation RM stands for \emph{RandomRooms}~\cite{rao:2021:randomrooms}, as the generation rules are adopted from this work.

\noindent
\textbf{Masked Autoencoder.}
We use the same setup to pre-train and fine-tune masked autoencoders as~\cite{li:2023:applying}. 
Specifically, a 3-layer transformer encoder~\cite{vaswani:2017:transformer} extracts features from input patches. 
A decoder with 2 layers is attached to the encoder for pre-training. 
All models are pre-trained for 120 epochs with a batch size of 64 and an AdamW optimizer~\cite{loshchilov:2017:decoupled}. 
The pre-trained encoder is evaluated for 3D object detection on the ScanNet~\cite{dai:2017:scannet_dataset} benchmark.  
3DETR~\cite{misra:2021:3detr} is used as the detection head and is fine-tuned for 1080 epochs with batch size 8. 
The mean average precision with a 3D-IoU threshold of \SI{25}{\percent} and \SI{50}{\percent} (\ie, AP25 and AP50) over 18 classes are evaluation matrices. 
We also fine-tune the pre-trained transformer for semantic segmentation on the S3DIS~\cite{armeni:2017:s3dis_dataset} dataset, an indoor dataset captured in 6 indoor areas. 
A segmentation head is connected to the pre-trained encoder. The entire network is fine-tuned for 300 epochs with batch size 12. Mean accuracy (mAcc) and mean IoU (mIoU) over 13 classes in Area 5 are reported as metrics. 

\noindent
\textbf{Contrastive Learning.}
We pre-trains a PointNet++~\cite{qi:2017:pointnet++} using contrastive learning, following previous works~\cite{zhang:2021:depthcontrast, rao:2021:randomrooms, yamada:2022:fractal_point, li:2022:closer}. 
The network architecture is adopted from~\cite{qi:2019:votenet}, which consists of 4 down-sampling modules~\cite{qi:2017:pointnet++} and 2 up-sampling modules~\cite{qi:2017:pointnet++}. 
The model is pre-trained using the well-known MoCoV2 pipeline~\cite{chen:2020:mocov2} based on the global instance-level correspondence. 
The PointNet++ is pre-trained for 120 epochs using an SGD optimizer, an initial learning rate of 0.01, a batch size of 12, and the cosine annealing schedule~\cite{loshchilov:2016:sgdr}. 
After pre-training, a VoteNet~\cite{qi:2019:votenet} is initialized with the pre-trained model and is fine-tuned for 3D object detection on the ScanNet~\cite{dai:2017:scannet_dataset} and the SUN RGB-D~\cite{song:2015:sunrgbd_dataset} benchmark. 
We report AP25 and AP50 as evaluation metrics. 

\subsection{Main Results}
\label{subsec:main}
The datasets in Table~\ref{tab:datagen_data} are used for pre-training models using masked autoencoder and contrastive learning. 

\noindent
\textbf{Masked Autoencoder.}
An MAE is pre-trained using the generated data, and the fine-tuning performance is reported in Table~\ref{tab:datagen_rm_mae}. 
Compared to results without pre-training~(row 1), all pre-trained models gain a significant improvement in object detection and semantic segmentation, showing the benefit of self-supervised pre-training.

Among the multi-view variants of randomly generated data~(row 3 to 6), using ShapeNet as an object set (\ie, RM-ShapeNet) delivers the best performance since ShapeNet is larger than other object sets (see Table~\ref{tab:datagen_data}). 
Meanwhile, RM-Fractal brings the slightest improvement. It supports our intuition that fractal point clouds as objects are non-optimal in data generation due to their unrealistic appearance and lack of surfaces. 
Interestingly, with the same object number, the RM-Harmonics dataset outperforms RM-ModelNet. 
The result is promising since it implies that randomly generated objects can replace CAD models for scene generation. 

However, most CAD models in ModelNet40 are ordinary objects in daily life, \eg, chairs and tables. At the same time, both datasets in downstream tasks are captured in indoor environments, \eg, living rooms and offices. 
One should expect ModelNet40 to perform better than spherical harmonics because of the similarity between pre-training and fine-tuning data. 
We believe the diversity of the object set might cause this contradiction. 
\begin{table}[t]
    \centering
	\setlength\tabcolsep{2.8pt}
	\begin{tabular}{lccc|cc}
		\toprule
		& & \multicolumn{2}{c|}{\textbf{Det.}} & \multicolumn{2}{c}{\textbf{Seg.}} \\
		\textbf{Pre-Training Data} & \textbf{View} & \textbf{AP25} & \textbf{AP50} & \textbf{mAcc} & \textbf{mIoU} \\
		\midrule
		N/A~\cite{li:2023:applying} & - & 61.6 & 38.8 & 66.4 & 60.0 \\
		ScanNet~\cite{li:2023:applying} & single & \textbf{64.1} & 43.0 & \textbf{74.7} & 67.6 \\
		\midrule
		RM-ShapeNet & multi & 63.8 & 42.6 & 74.6 & \textbf{67.8} \\
		RM-ModelNet & multi & 63.5 & 42.0 & 73.6 & 67.0 \\
		RM-Fractal & multi & 62.8 & 40.4 & 70.2 & 64.1 \\
		RM-Harmonics & multi & 63.8 & \textbf{43.2} & 74.1 & 67.1 \\   
		\midrule
		RM-ShapeNet & single & 63.8 & 42.9 & 73.8 & 67.2 \\
		RM-ModelNet & single & 62.5 & 41.6 & 73.5 & 66.3 \\
		RM-Harmonics & single & 63.5 & 42.1 & 73.5 & 67.1 \\  
		\bottomrule
	\end{tabular}
	\caption{Downstream task performance of an MAE pre-trained on different datasets. Detection results are on the ScanNet detection benchmark. Segmentation results are on the S3DIS dataset. The second column ``View'' refers to the pre-training data. All metrics are in percentage. }
	\label{tab:datagen_rm_mae}
\end{table}
Although ModelNet40 contains 40 classes, the appearance of the objects within each class is similar. Also, the class distribution in ModelNet40 is long-tailed~\cite{wu:2015:modelnet_dataset}, which might also have a negative impact~\cite{tian:2021:divide}. 
On the other hand, spherical harmonics are generated by uniformly sampling the coefficients, which results in a more diverse and balanced object set. 

Furthermore, Table~\ref{tab:datagen_rm_mae} compares the results of generated synthetic data with the real-world ScanNet dataset (row 2). 
With the same number of point clouds, the performance of RM-ShapeNet and RM-Harmonics is close to real-world data, demonstrating the effectiveness of randomized scene generation in self-supervised pre-training. 

With generated single-view data (\ie, depth maps), a similar trend can also be observed: RM-ShapeNet performs the best, while RM-Harmonics is more effective than RM-ModelNet. 
Data generation with fractal points is not performed since ray-casting relies on mesh data. However, compared to their corresponding multi-view variants, all single-view datasets perform worse with MAE. 

\noindent
\textbf{Contrastive Learning.}
\label{subsubsec:datagen_exp_cl}
For contrastive learning, a PointNet++ is pre-trained, and a VoteNet is fine-tuned using it as the backbone. 
Object detection results on ScanNet and SUN RGB-D are reported in Table~\ref{tab:datagen_rm_contrast}. 
Generally, pre-trained models achieve better detection quality than the baseline without pre-training on both benchmarks. 
For the multi-view datasets (row 3 to 6), RM-ShapeNet and RM-Harmonics show on-par performance. RM-ModelNet performs worse than these two datasets, while RM-Fractal brings the smallest improvement in pre-training. 
Compared to real-world data from ScanNet, randomly generated data shows significantly worse results on the SUN RGB-D dataset, while the gap disappears on the ScanNet detection benchmark. 

\begin{table}[t]
	\centering
	\setlength\tabcolsep{2pt}
	\begin{tabular}{lccc|cc}
		\toprule
		& & \multicolumn{2}{c|}{\textbf{SUN RGB-D}} & \multicolumn{2}{c}{\textbf{ScanNet}} \\
		\textbf{Pre-Training Data} & \textbf{View} & \textbf{AP25} & \textbf{AP50} & \textbf{AP25} & \textbf{AP50} \\
		\midrule
		N/A~\cite{qi:2019:votenet} & - & 58.4 & 33.3 & 60.0 & 37.6 \\ 
		ScanNet & single & \textbf{60.8} & \textbf{35.7} & 62.2 & 39.0 \\
		\midrule
		RM-ShapeNet & multi & 59.8 & 34.1 & 62.7 & 37.9 \\
		RM-ModelNet & multi & 59.3 & 33.7 & 62.3 & 37.3 \\
		RM-Fractal & multi & 59.1 & 33.4 & 61.3 & 38.6 \\
		RM-Harmonics & multi & 59.6 & \textbf{35.7} & 62.6 & 39.7 \\
		\midrule
		RM-ShapeNet & single & 58.9 & 33.0 & \textbf{64.1} & \textbf{40.0} \\
		RM-ModelNet & single & 57.9 & 32.4 & 63.4 & 39.8 \\
		RM-Harmonics & single & 58.6 & 33.1 & 63.5 & 39.8 \\
		\midrule
		RandomRooms~\cite{rao:2021:randomrooms} & multi & 59.2 & - & 61.3 & - \\
		PC-FractalDB~\cite{yamada:2022:fractal_point} & multi & 59.4 & 33.9 & 61.9 & 38.3 \\
		\bottomrule
	\end{tabular}
	\caption{Fine-tuning results of contrastive learning using generated data. Evaluation for 3D object detection on SUN RGB-D and ScanNet benchmark. All metrics in percentage. Absent values are not reported in the original publication. }
	\label{tab:datagen_rm_contrast}
\end{table}

The relative performance of single-view variants of generated data is similar to the multi-view ones, \eg, RM-ShapeNet $>$ RM-Harmonic $>$ RM-ModelNet. 
However, the detection results on the SUN RGB-D dataset are worse compared to using multi-view data. 
On the contrary, single-view datasets show significantly better results on the ScanNet benchmark. 
The synthetic data even outperform the real-world baseline by a clear margin. 
This observation is counter-intuitive. Note that SUN RGB-D consists of single-view point clouds, whereas the ScanNet benchmark applies point clouds reconstructed from multiple views. 
One should expect the models pre-trained using single-view data to perform worse on the ScanNet benchmark due to the domain gap. The reason for this observation is not fully understood. However, it must be correlated with the pre-training schema since it is not observed with MAE. 

Table~\ref{tab:datagen_rm_contrast} also compares the generated datasets with two previous works, \ie, RandomRooms~\cite{rao:2021:randomrooms} and PC-FractalDB~\cite{yamada:2022:fractal_point}, each applies a training strategy specialized for object detection (see Section~\ref{subsec:revisit}). 
Both methods use VoteNet as the detector and apply PointNet++ as the backbone. 
The model pre-trained with RM-Fractal achieves similar performance as PC-FractalDB~\cite{yamada:2022:fractal_point}, which also generates scenes using fractal point clouds. 
However, RM-Harmonics (multi-view) outperforms both previous works by a clear margin. It shows that the choice of object sets has a larger impact than pre-training methods. Also, it implies that a task-specific design is unnecessary in pre-training and general purpose methods (\eg, MAE and MoCoV2) are sufficient. 

\noindent
\textbf{Discussions.}
The experimental results in Table~\ref{tab:datagen_rm_mae} and \ref{tab:datagen_rm_contrast} can be summarized as follows: 

\begin{enumerate}
	\item Randomized scene generation is effective for pre-training 3D models with self-supervision. This approach generalizes well across different pre-training methods and downstream tasks. Also, its performance is comparable with real-world data. 
	\item Formula-driven spherical harmonics perform significantly better than fractal point clouds and achieve on-par results as hand-crafted CAD models. 
	\item The diversity of objects impacts the effect of pre-training. An object set with a larger diversity is generally beneficial. 
	\item Single-view and multi-view variants of generated data bring different results. However, the impact is inconsistent across different pre-training methods and fine-tuning datasets. 
\end{enumerate}

The observation that randomly generated data have a similar effect as real-world ones in self-supervised pre-training is interesting since the synthetic data do not look photo-realistic (see Figure~\ref{fig:scenes_pcd} and \ref{fig:depth}), especially when formula-driven objects are applied. 
We believe it is because neural networks perform different tasks in self-supervised pre-training and supervised fine-tuning (\ie, pre-training and downstream tasks are decoupled). 
Therefore, the performance in fine-tuning is not sensitive to the domain gap between the pre-training and fine-tuning data. 
On the other hand, some useful features are less affected by the domain gap, \eg, features describing corners and edges or distinguishing objects from backgrounds. 
Nevertheless, a large gap can still harm the performance, \eg, fractal point clouds perform consistently worse than spherical harmonics in our experiments. 

\subsection{Additional Experiments}

\noindent
\textbf{Comparison with other Pre-Training Methods.}
Table~\ref{tab:datagen_det} compares the fine-tuning performance of models pre-trained on RM-Harmonics with other state-of-the-art methods. 
All methods indicated with ``real'' are pre-trained on real-world data from ScanNet, while DepthContrast ($\times3$)~in Table~\ref{tab:datagen_det} additionally uses data from the larger Redwood indoor RGB-D scan dataset~\cite{park:2017:colored}. 
All methods in the upper half of Table~\ref{tab:datagen_det} fine-tune a VoteNet with a pre-trained PointNet++ as the backbone, while methods in the lower half use a 3DETR with a pre-trained transformer. 

Compared with methods using real-world data for pre-training, the proposed RM-Harmonics achieves competitive performance. 
However, pre-training using RM-Harmonics is more efficient since the data are randomly generated and require neither real-world 3D scans nor hand-crafted CAD models. 

\begin{table}[htb]
	\centering
	\setlength\tabcolsep{1.8pt}
	\begin{tabular}{lcccc}
		\toprule
		\textbf{Methods} & \textbf{Data} & \textbf{Model} & \textbf{AP25} & \textbf{AP50} \\
		\midrule
		VoteNet~\cite{qi:2019:votenet} & N/A & P & 60.0 & 37.6 \\
		Hou~\etal~\cite{hou:2021:exploring_efficient} & real & P & - & 39.3 \\
		DepthContrast ($\times1$)~\cite{zhang:2021:depthcontrast} & real & P & 61.3 & - \\
		DepthContrast ($\times3$)~\cite{zhang:2021:depthcontrast} & real & P & 64.0 & 42.9 \\
		DPCo~\cite{li:2022:closer} & real & P & \textbf{64.2} & 41.5 \\
		RandomRooms~\cite{rao:2021:randomrooms} & syn. & P & 61.3 & - \\
		PC-FractalDB~\cite{yamada:2022:fractal_point} & syn. & P & 61.9 & 38.3 \\
		RM-Harmonics (multi-view) & syn. & P & 62.6 & 39.7 \\
		RM-Harmonics (single-view) & syn. & P & 63.5 & 39.8 \\
		\midrule
		3DETR~\cite{misra:2021:3detr} & N/A & T & 62.1 & 37.9 \\
		MaskPoint (L3)~\cite{liu:2022:masked_discrimi} & real & T & 63.4 & 40.6 \\
		MaskPoint (L12)~\cite{liu:2022:masked_discrimi} & real & T & \textbf{64.2} & 42.1 \\
		Plain Transformer~\cite{li:2023:applying} & real & T & 64.1 & 43.0 \\
		RM-Harmonics (multi-view) & syn. & T & 63.8 & \textbf{43.2} \\
		RM-Harmonics (single-view) & syn. & T & 63.5 & 42.1 \\
		\bottomrule
	\end{tabular}
	\caption{Object detection results on the ScanNet detection benchmark. Data: type of pre-training data (real-world, synthetic or none). Model: the architecture of backbones. P: PointNet++, T: transformer. All reported values are in percentage. }
	\label{tab:datagen_det}
\end{table}

\noindent
\textbf{Pseudo Color in Pre-Training.}
Since simulation and rendering are not involved, the scenes generated in this work do not contain color information. 
However, color channels are crucial in some downstream tasks, \eg, semantic segmentation. To transfer models trained on colorless point clouds to such tasks, we use pseudo colors and apply data augmentation to the color channels in pre-training. Specifically, all color channels are set to a constant value~(\ie, 0.5), and random color jitter is applied to each point. 
Also, color values of all points are simultaneously set to 0 with the probability 0.5~(\ie, color drop out). 

\begin{table}[b]
	\centering
        \setlength\tabcolsep{3.8pt}
	\begin{tabular}{cccc}
		\toprule
		\textbf{Pre-Train Data} & \textbf{Colors} & \textbf{mAcc} (\%) & \textbf{mIoU} (\%) \\
		\midrule
		N/A~\cite{li:2023:applying} & N/A & 66.4 & 60.0 \\
            \midrule
		RM-Harmonics & const. & 69.8 & 62.8 \\
		RM-Harmonics & const. + DA & 74.1 & 67.1 \\
		\midrule
		ScanNet & const. & 69.8 & 63.6 \\
		ScanNet & const. + DA & 73.1 & 67.0 \\
		ScanNet & real & \textbf{74.7} & \textbf{67.6} \\
		\bottomrule
	\end{tabular}
	\caption{Impact of colors in pre-training using MAE. Fine-tuned for semantic segmentation on the S3DIS dataset and evaluated in Area 5. The first two columns refer to the pre-training data. Const.: constant RGB values for all points. DA: data augmentation on color channels. }
	\label{tab:datagen_color}
\end{table}

As shown in Table~\ref{tab:datagen_color}, when no data augmentation (\ie, jitter and drop out) is applied to color values, the constant pseudo color leads to significantly worse results with RM-Harmonics. 
Table~\ref{tab:datagen_color} also shows the results of models pre-trained on ScanNet to compare the performance of pseudo colors and real-world colors. 
Without data augmentation on pseudo colors, the performance of pseudo colors is much lower than real-world colors. However, applying data augmentation shrinks the gap significantly. 
Interestingly, when both use pseudo colors, RM-Harmonics slightly outperforms ScanNet in pre-training. 
The results in Table~\ref{tab:datagen_color} demonstrate that although the generated synthetic data are colorless, they can still be applied to pre-train models requiring colors. 
Additionally, data augmentation is essential for improving performance. 
We hypothesize that models learn color invariance via data augmentation and focus on the geometric information in the pre-training. In contrast, color-dependent information is learned in fine-tuning where real-world colors are available.

\begin{figure}[htb]
    \centering
		\begin{tikzpicture}[scale=0.9]
			\begin{axis}[
				xlabel={Used Data (\%)},
				ylabel={AP25 (\%)},
				ymin = 0,
				ymax = 67.5,
				xmin = 0,
				xmax = 100,
				xtick = {5, 10, 20, 50, 100},
				ytick = {10, 20, 30, 40, 50, 60},
    		      y=0.05cm,
				legend style={at={(0.73,0.50)},anchor=north,legend columns=1},
				ymajorgrids=true,
				grid style=dashed,
				line width=1.2pt,
				mark size=1.5pt
				]
				
    	\definecolor{kitgreen}{RGB}{0,170,130}
		\definecolor{kitblue}{RGB}{70,50,170}
		\definecolor{kitred}{RGB}{190,14,15}	
				
				\addplot[color=kitred, mark=*,ultra thick]
				coordinates {
					(5, 3.57)(10, 7.36)(20, 31.73)(50, 52.59)(100, 61.6)
				};
				
				\addplot[color=kitblue, mark=*,ultra thick]
				coordinates {
					(5, 26.61)(10, 37.66)(20, 47.89) (50, 57.68)(100, 64.0)
				};
				
				\addplot[color=kitgreen, mark=*,ultra thick]
				coordinates {
					(5, 22.33)(10, 34.44)(20, 46.36) (50, 57.30)(100, 63.8)
				};
			
				\legend{From scratch, ScanNet, RM-Harmonics}
				
			\end{axis}
		\end{tikzpicture}
	\caption{Detection results of MAE with reduced fine-tuning data and labels.}
        \label{fig:datagen_label_eff}
\end{figure}
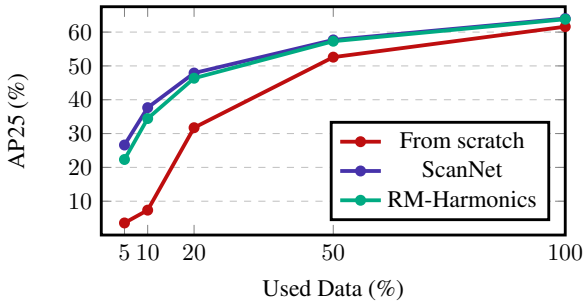

\noindent
\textbf{Reduced Fine-Tuning Data.}
Pre-training is especially useful when fine-tuning data are scarce. To verify if our method applies to this scenario, we pre-train an MAE on the generated RM-Harmonics dataset. Then, the encoder is fine-tuned on ScanNet for object detection using \SI{5}{\percent}, \SI{10}{\percent}, \SI{20}{\percent}, \SI{50}{\percent}, and \SI{100}{\percent} data, respectively. 
Figure~\ref{fig:datagen_label_eff} demonstrates that pre-training with RM-Harmonics has a similar effect as ScanNet: the improvement from pre-training is more significant when fewer data are used in fine-tuning. 
However, when fewer annotated samples (\eg, \SI{5}{\percent} and \SI{10}{\percent}) are available, real-world ScanNet performs better than randomly generated RM-Harmonics. The performance gap is closed when the detector is fine-tuned with more annotated data (\eg, more than \SI{20}{\percent}). 
This observation demonstrates that the RM-Harmonics dataset indeed has a domain gap. 
However, models pre-trained on this dataset can quickly adapt to the real-world dataset during fine-tuning. 

\noindent
\textbf{Pre-Pre-Training.}
Section~\ref{subsec:main} demonstrates that randomized synthetic scenes can replace real-world captures for pre-training. 
In this experiment, we study if synthetic data are still beneficial when real-world pre-training data are available. 
To this end, we apply a two-stage pipeline using an MAE. 
Specifically, the RM-Harmonics dataset is used for the first stage (\ie, pre-pre-training). 
Then, the MAE is additionally trained using real-world data from ScanNet (\ie, pre-training). Both stages use the setup in Section~\ref{subsec:setup}. 
The fine-tuning results of pre-trained models are shown in Table~\ref{tab:two_round}. 
Compared to pre-training exclusively on ScanNet, this two-stage method (the last row) improves the AP25 and AP50 by \SI{0.1}{\percent} and \SI{0.6}{\percent}, respectively. 

We also test other choices, \eg, doubling the pre-training samples from ScanNet (indicated by $\times$2 in Table~\ref{tab:two_round}), mixing ScanNet and RM-Harmonics for pre-training in one stage (mixture), and exchanging the order of ScanNet and RM-Harmonics in the two-stage method. 
\begin{table}[t]
    \centering
    \setlength\tabcolsep{3.5pt}
    \begin{tabular}{cccc}
    \toprule
         \textbf{Pre-Pre-Train} & \textbf{Pre-Train} & \textbf{AP25} (\%) & \textbf{AP50} (\%) \\
    \midrule
         - & - & 61.6 & 38.8 \\
         - & RM-Harmonics & 63.8 & 43.2 \\
         - & ScanNet & 64.1 & 43.0 \\
         \midrule
         - & ScanNet ($\times$2)  & 63.9 & 43.0 \\
         - & mixture & 62.9 & 42.0 \\
         ScanNet & RM-Harmonics & 63.6 & 42.1 \\
         RM-Harmonics & ScanNet & \textbf{64.2} & \textbf{43.6} \\
    \bottomrule
    \end{tabular}
    \caption{Fine-tuning results of two-stage pre-training. Evaluated for 3D object detection on the ScanNet dataset. The last four rows use an increased amount of data compared to the default setup.}
    \label{tab:two_round}
\end{table}
However, these three variants do not improve the performance compared to the real-world baseline (the third row). 
Sampling more data from ScanNet is not effective because the dataset is captured in 1500 indoor rooms. The diversity of the data is bounded. 
The mixture of two datasets does not perform well since the model has to learn two different distributions simultaneously, which has a negative impact. 
Moreover, the ``RM-Harmonics - ScanNet'' variant of two-stage pre-training is more effective than the ``ScanNet - RM-Harmonics'' variant since the former allows the model to adapt to the real-world data distribution before fine-tuning. 

After all, this experiment demonstrates another usage of randomized 3D scene generation: the synthetic data can enrich pre-training data by supplementing real-world data (instead of replacing them). 


\section{Conclusion and Future Works}
This work rethinks randomized 3D scene generation for generalizable self-supervised pre-training. 
We demonstrate that the generated data generalize well with different pre-training methods and achieves competitive results in multiple downstream tasks. 
Furthermore, we show that formula-driven spherical harmonics as objects perform similarly to CAD models and even real-world data. 

The simplicity of randomized 3D scene generation is especially promising since it allows pre-training strong models at a low cost.
We believe this technique works with objects other than 
CAD models and spherical harmonics. 
Generating objects using other parametric shapes or generative deep learning models can be an interesting topic for future works. 
We hope this work could inspire more research interest in randomized 3D scene generation. 

{
    \small
    \bibliographystyle{ieeenat_fullname}
    \bibliography{main}

\begin{thebibliography}{49}
\providecommand{\natexlab}[1]{#1}
\providecommand{\url}[1]{\texttt{#1}}
\expandafter\ifx\csname urlstyle\endcsname\relax
  \providecommand{\doi}[1]{doi: #1}\else
  \providecommand{\doi}{doi: \begingroup \urlstyle{rm}\Url}\fi

\bibitem[Armeni et~al.(2017)Armeni, Sax, Zamir, and
  Savarese]{armeni:2017:s3dis_dataset}
Iro Armeni, Sasha Sax, Amir~Roshan Zamir, and Silvio Savarese.
\newblock Joint {2D}-{3D}-semantic data for indoor scene understanding.
\newblock \emph{CoRR}, abs/1702.01105, 2017.

\bibitem[Chang et~al.(2015)Chang, Funkhouser, Guibas, Hanrahan, Huang, Li,
  Savarese, Savva, Song, Su, et~al.]{chang:2015:shapenet_dataset}
Angel~X. Chang, Thomas Funkhouser, Leonidas Guibas, Pat Hanrahan, Qixing Huang,
  Zimo Li, Silvio Savarese, Manolis Savva, Shuran Song, Hao Su, et~al.
\newblock {ShapeNet}: An information-rich {3D} model repository.
\newblock \emph{arXiv preprint arXiv:1512.03012}, 2015.

\bibitem[Chen et~al.(2020)Chen, Fan, Girshick, and He]{chen:2020:mocov2}
Xinlei Chen, Haoqi Fan, Ross Girshick, and Kaiming He.
\newblock Improved baselines with momentum contrastive learning.
\newblock \emph{arXiv preprint arXiv:2003.04297}, 2020.

\bibitem[Dai et~al.(2017)Dai, Chang, Savva, Halber, Funkhouser, and
  Nie{\ss}ner]{dai:2017:scannet_dataset}
Angela Dai, Angel~X. Chang, Manolis Savva, Maciej Halber, Thomas Funkhouser,
  and Matthias Nie{\ss}ner.
\newblock {ScanNet}: Richly-annotated {3D} reconstructions of indoor scenes.
\newblock In \emph{Proceedings of the IEEE conference on computer vision and
  pattern recognition}, pages 5828--5839, 2017.

\bibitem[Deschaud et~al.(2021)Deschaud, Duque, Richa, Velasco-Forero,
  Marcotegui, and Goulette]{deschaud:2021:paris_carla}
Jean-Emmanuel Deschaud, David Duque, Jean~Pierre Richa, Santiago
  Velasco-Forero, Beatriz Marcotegui, and Fran{\c{c}}ois Goulette.
\newblock Paris-carla-3d: A real and synthetic outdoor point cloud dataset for
  challenging tasks in {3D} mapping.
\newblock \emph{Remote Sensing}, 13\penalty0 (22):\penalty0 4713, 2021.

\bibitem[Dosovitskiy et~al.(2017)Dosovitskiy, Ros, Codevilla, Lopez, and
  Koltun]{Dosovitskiy:17:carla_simulator}
Alexey Dosovitskiy, German Ros, Felipe Codevilla, Antonio Lopez, and Vladlen
  Koltun.
\newblock {CARLA}: {An} open urban driving simulator.
\newblock In \emph{Proceedings of the 1st Annual Conference on Robot Learning},
  pages 1--16, 2017.

\bibitem[Dosovitskiy et~al.(2020)Dosovitskiy, Beyer, Kolesnikov, Weissenborn,
  Zhai, Unterthiner, Dehghani, Minderer, Heigold, Gelly,
  et~al.]{dosovitskiy:2020:vit}
Alexey Dosovitskiy, Lucas Beyer, Alexander Kolesnikov, Dirk Weissenborn,
  Xiaohua Zhai, Thomas Unterthiner, Mostafa Dehghani, Matthias Minderer, Georg
  Heigold, Sylvain Gelly, et~al.
\newblock An image is worth 16x16 words: Transformers for image recognition at
  scale.
\newblock \emph{arXiv preprint arXiv:2010.11929}, 2020.

\bibitem[Fu et~al.(2022)Fu, Gao, Liu, Zhang, Qiao, and Wang]{fu:2022:pos_bert}
Kexue Fu, Peng Gao, ShaoLei Liu, Renrui Zhang, Yu Qiao, and Manning Wang.
\newblock {POS-BERT}: Point cloud one-stage {BERT} pre-training.
\newblock \emph{arXiv preprint arXiv:2204.00989}, 2022.

\bibitem[Griffiths and Boehm(2019)]{griffiths:2019:synthcity}
David Griffiths and Jan Boehm.
\newblock {SynthCity}: A large scale synthetic point cloud.
\newblock \emph{arXiv preprint arXiv:1907.04758}, 2019.

\bibitem[Gschwandtner et~al.(2011)Gschwandtner, Kwitt, Uhl, and
  Pree]{gschwandtner:2011:blensor}
Michael Gschwandtner, Roland Kwitt, Andreas Uhl, and Wolfgang Pree.
\newblock {BlenSor}: {Blender} sensor simulation toolbox.
\newblock In \emph{Advances in Visual Computing: 7th International Symposium,
  ISVC 2011, Las Vegas, NV, USA, September 26-28, 2011. Proceedings, Part II
  7}, pages 199--208. Springer, 2011.

\bibitem[He et~al.(2022)He, Chen, Xie, Li, Doll{\'a}r, and
  Girshick]{he:2022:mae}
Kaiming He, Xinlei Chen, Saining Xie, Yanghao Li, Piotr Doll{\'a}r, and Ross
  Girshick.
\newblock Masked autoencoders are scalable vision learners.
\newblock In \emph{Proceedings of the IEEE/CVF Conference on Computer Vision
  and Pattern Recognition}, pages 16000--16009, 2022.

\bibitem[Hobolth et~al.(2003)Hobolth, Pedersen, and
  Jensen]{hobolth:2003:continuous}
Asger Hobolth, Jan Pedersen, and Eva B~Vedel Jensen.
\newblock A continuous parametric shape model.
\newblock \emph{Annals of the Institute of Statistical Mathematics},
  55:\penalty0 227--242, 2003.

\bibitem[Hou et~al.(2021)Hou, Graham, Nie{\ss}ner, and
  Xie]{hou:2021:exploring_efficient}
Ji Hou, Benjamin Graham, Matthias Nie{\ss}ner, and Saining Xie.
\newblock Exploring data-efficient {3D} scene understanding with contrastive
  scene contexts.
\newblock In \emph{Proceedings of the IEEE/CVF Conference on Computer Vision
  and Pattern Recognition}, pages 15587--15597, 2021.

\bibitem[Hurl et~al.(2019)Hurl, Czarnecki, and
  Waslander]{hurl:2019:presil_dataset}
Braden Hurl, Krzysztof Czarnecki, and Steven Waslander.
\newblock Precise synthetic image and lidar (presil) dataset for autonomous
  vehicle perception.
\newblock In \emph{2019 IEEE Intelligent Vehicles Symposium (IV)}, pages
  2522--2529. IEEE, 2019.

\bibitem[Jun and Nichol(2023)]{jun:2023:shap}
Heewoo Jun and Alex Nichol.
\newblock {Shap-E}: Generating conditional 3d implicit functions.
\newblock \emph{arXiv preprint arXiv:2305.02463}, 2023.

\bibitem[Kazhdan et~al.(2003)Kazhdan, Funkhouser, and
  Rusinkiewicz]{kazhdan:2003:rotation}
Michael Kazhdan, Thomas Funkhouser, and Szymon Rusinkiewicz.
\newblock Rotation invariant spherical harmonic representation of {3D} shape
  descriptors.
\newblock In \emph{Symposium on geometry processing}, pages 156--164, 2003.

\bibitem[Li and Heizmann(2022)]{li:2022:closer}
Lanxiao Li and Michael Heizmann.
\newblock A closer look at invariances in self-supervised pre-training for {3D}
  vision.
\newblock In \emph{Computer Vision--ECCV 2022: 17th European Conference, Tel
  Aviv, Israel, October 23--27, 2022, Proceedings, Part XXX}, pages 656--673.
  Springer, 2022.

\bibitem[Li and Heizmann(2023)]{li:2023:applying}
Lanxiao Li and Michael Heizmann.
\newblock Applying plain transformers to real-world point clouds.
\newblock \emph{arXiv preprint arXiv:2303.00086}, 2023.

\bibitem[Liu et~al.(2022)Liu, Cai, and Lee]{liu:2022:masked_discrimi}
Haotian Liu, Mu Cai, and Yong~Jae Lee.
\newblock Masked discrimination for self-supervised learning on point clouds.
\newblock \emph{Proceedings of the European Conference on Computer Vision
  (ECCV)}, 2022.

\bibitem[Loshchilov and Hutter(2016)]{loshchilov:2016:sgdr}
Ilya Loshchilov and Frank Hutter.
\newblock Sgdr: Stochastic gradient descent with warm restarts.
\newblock \emph{arXiv preprint arXiv:1608.03983}, 2016.

\bibitem[Loshchilov and Hutter(2017)]{loshchilov:2017:decoupled}
Ilya Loshchilov and Frank Hutter.
\newblock Decoupled weight decay regularization.
\newblock \emph{arXiv preprint arXiv:1711.05101}, 2017.

\bibitem[Misra et~al.(2021)Misra, Girdhar, and Joulin]{misra:2021:3detr}
Ishan Misra, Rohit Girdhar, and Armand Joulin.
\newblock An end-to-end transformer model for {3D} object detection.
\newblock In \emph{Proceedings of the IEEE/CVF International Conference on
  Computer Vision}, pages 2906--2917, 2021.

\bibitem[Monedero(2000)]{monedero:2000:parametric}
Javier Monedero.
\newblock Parametric design: a review and some experiences.
\newblock \emph{Automation in construction}, 9\penalty0 (4):\penalty0 369--377,
  2000.

\bibitem[M{\"u}ller(2006)]{muller:2006:spherical}
Claus M{\"u}ller.
\newblock \emph{Spherical harmonics}.
\newblock Springer, 2006.

\bibitem[Nichol et~al.(2022)Nichol, Jun, Dhariwal, Mishkin, and
  Chen]{nichol:2022:pointe}
Alex Nichol, Heewoo Jun, Prafulla Dhariwal, Pamela Mishkin, and Mark Chen.
\newblock {Point-E}: a system for generating 3d point clouds from complex
  prompts.
\newblock \emph{arXiv preprint arXiv:2212.08751}, 2022.

\bibitem[Pang et~al.(2022)Pang, Wang, Tay, Liu, Tian, and
  Yuan]{pang:2022:point_mae}
Yatian Pang, Wenxiao Wang, Francis~EH Tay, Wei Liu, Yonghong Tian, and Li Yuan.
\newblock Masked autoencoders for point cloud self-supervised learning.
\newblock \emph{ECCV}, 2022.

\bibitem[Park et~al.(2017)Park, Zhou, and Koltun]{park:2017:colored}
Jaesik Park, Qian-Yi Zhou, and Vladlen Koltun.
\newblock Colored point cloud registration revisited.
\newblock In \emph{Proceedings of the IEEE international conference on computer
  vision}, pages 143--152, 2017.

\bibitem[Qi et~al.(2017)Qi, Yi, Su, and Guibas]{qi:2017:pointnet++}
Charles~R. Qi, Li Yi, Hao Su, and Leonidas~J. Guibas.
\newblock {PointNet++}: Deep hierarchical feature learning on point sets in a
  metric space.
\newblock \emph{Advances in neural information processing systems}, 30, 2017.

\bibitem[Qi et~al.(2019)Qi, Litany, He, and Guibas]{qi:2019:votenet}
Charles~R. Qi, Or Litany, Kaiming He, and Leonidas~J. Guibas.
\newblock Deep {Hough} voting for {3D} object detection in point clouds.
\newblock In \emph{proceedings of the IEEE/CVF International Conference on
  Computer Vision}, pages 9277--9286, 2019.

\bibitem[Qian et~al.(2022)Qian, Zhang, Hamdi, and Ghanem]{qian:2022:pix4point}
Guocheng Qian, Xingdi Zhang, Abdullah Hamdi, and Bernard Ghanem.
\newblock {Pix4Point}: Image pretrained transformers for {3D} point cloud
  understanding.
\newblock \emph{arXiv preprint arXiv:2208.12259}, 2022.

\bibitem[Rao et~al.(2021)Rao, Liu, Wei, Lu, Hsieh, and
  Zhou]{rao:2021:randomrooms}
Yongming Rao, Benlin Liu, Yi Wei, Jiwen Lu, Cho-Jui Hsieh, and Jie Zhou.
\newblock Randomrooms: unsupervised pre-training from synthetic shapes and
  randomized layouts for {3D} object detection.
\newblock In \emph{Proceedings of the IEEE/CVF International Conference on
  Computer Vision}, pages 3283--3292, 2021.

\bibitem[Roth(1982)]{roth:1982:raycast}
Scott~D Roth.
\newblock Ray casting for modeling solids.
\newblock \emph{Computer graphics and image processing}, 18\penalty0
  (2):\penalty0 109--144, 1982.

\bibitem[Saupe and Vrani{\'c}(2001)]{saupe:2001:3d}
Dietmar Saupe and Dejan~V Vrani{\'c}.
\newblock {3D} model retrieval with spherical harmonics and moments.
\newblock In \emph{Pattern Recognition: 23rd DAGM Symposium Munich, Germany,
  September 12--14, 2001 Proceedings 23}, pages 392--397. Springer, 2001.

\bibitem[Schulz et~al.(2017)Schulz, Shamir, Baran, Levin, Sitthi-Amorn, and
  Matusik]{schulz:2017:retrieval}
Adriana Schulz, Ariel Shamir, Ilya Baran, David~IW Levin, Pitchaya
  Sitthi-Amorn, and Wojciech Matusik.
\newblock Retrieval on parametric shape collections.
\newblock \emph{ACM Transactions on Graphics (TOG)}, 36\penalty0 (1):\penalty0
  1--14, 2017.

\bibitem[Smirnov et~al.(2020)Smirnov, Fisher, Kim, Zhang, and
  Solomon]{smirnov:2020:deep}
Dmitriy Smirnov, Matthew Fisher, Vladimir~G Kim, Richard Zhang, and Justin
  Solomon.
\newblock Deep parametric shape predictions using distance fields.
\newblock In \emph{Proceedings of the IEEE/CVF Conference on Computer Vision
  and Pattern Recognition}, pages 561--570, 2020.

\bibitem[Song et~al.(2015)Song, Lichtenberg, and
  Xiao]{song:2015:sunrgbd_dataset}
Shuran Song, Samuel~P. Lichtenberg, and Jianxiong Xiao.
\newblock {SUN RGB-D}: A {RGB-D} scene understanding benchmark suite.
\newblock In \emph{Proceedings of the IEEE conference on computer vision and
  pattern recognition}, pages 567--576, 2015.

\bibitem[Tian et~al.(2021)Tian, Henaff, and van~den Oord]{tian:2021:divide}
Yonglong Tian, Olivier~J Henaff, and A{\"a}ron van~den Oord.
\newblock Divide and contrast: Self-supervised learning from uncurated data.
\newblock In \emph{Proceedings of the IEEE/CVF International Conference on
  Computer Vision}, pages 10063--10074, 2021.

\bibitem[Vaswani et~al.(2017)Vaswani, Shazeer, Parmar, Uszkoreit, Jones, Gomez,
  Kaiser, and Polosukhin]{vaswani:2017:transformer}
Ashish Vaswani, Noam Shazeer, Niki Parmar, Jakob Uszkoreit, Llion Jones,
  Aidan~N. Gomez, {\L}ukasz Kaiser, and Illia Polosukhin.
\newblock Attention is all you need.
\newblock \emph{Advances in neural information processing systems}, 30, 2017.

\bibitem[Wang et~al.(2021)Wang, Liu, Yue, Lasenby, and
  Kusner]{wang:2021:occlusion}
Hanchen Wang, Qi Liu, Xiangyu Yue, Joan Lasenby, and Matt~J. Kusner.
\newblock Unsupervised point cloud pre-training via occlusion completion.
\newblock In \emph{Proceedings of the IEEE/CVF international conference on
  computer vision}, pages 9782--9792, 2021.

\bibitem[Wu et~al.(2018)Wu, Wan, Yue, and Keutzer]{wu:2018:squeezeseg}
Bichen Wu, Alvin Wan, Xiangyu Yue, and Kurt Keutzer.
\newblock Squeezeseg: Convolutional neural nets with recurrent crf for
  real-time road-object segmentation from {3D} lidar point cloud.
\newblock In \emph{2018 IEEE international conference on robotics and
  automation (ICRA)}, pages 1887--1893. IEEE, 2018.

\bibitem[Wu et~al.(2015)Wu, Song, Khosla, Yu, Zhang, Tang, and
  Xiao]{wu:2015:modelnet_dataset}
Zhirong Wu, Shuran Song, Aditya Khosla, Fisher Yu, Linguang Zhang, Xiaoou Tang,
  and Jianxiong Xiao.
\newblock {3D ShapeNets}: A deep representation for volumetric shapes.
\newblock In \emph{Proceedings of the IEEE conference on computer vision and
  pattern recognition}, pages 1912--1920, 2015.

\bibitem[Xiao et~al.(2022)Xiao, Huang, Guan, Zhan, and
  Lu]{xiao:2022:transfer_from_synthetic}
Aoran Xiao, Jiaxing Huang, Dayan Guan, Fangneng Zhan, and Shijian Lu.
\newblock Transfer learning from synthetic to real lidar point cloud for
  semantic segmentation.
\newblock In \emph{Proceedings of the AAAI Conference on Artificial
  Intelligence}, pages 2795--2803, 2022.

\bibitem[Xie et~al.(2020)Xie, Gu, Guo, Qi, Guibas, and
  Litany]{xie:2020:pointcontrast}
Saining Xie, Jiatao Gu, Demi Guo, Charles~R Qi, Leonidas Guibas, and Or Litany.
\newblock {PointContrast}: Unsupervised pre-training for {3D} point cloud
  understanding.
\newblock In \emph{Computer Vision--ECCV 2020: 16th European Conference,
  Glasgow, UK, August 23--28, 2020, Proceedings, Part III 16}, pages 574--591.
  Springer, 2020.

\bibitem[Xu et~al.(2022)Xu, Yang, Zhai, Wu, Yue, Zhan, Vajda, Keutzer, and
  Tomizuka]{xu:2022:image2point}
Chenfeng Xu, Shijia Yang, Bohan Zhai, Bichen Wu, Xiangyu Yue, Wei Zhan, Peter
  Vajda, Kurt Keutzer, and Masayoshi Tomizuka.
\newblock {Image2point}: {3D} point-cloud understanding with pretrained 2d
  convnets.
\newblock \emph{Proceedings of the European Conference on Computer Vision
  (ECCV)}, 2022.

\bibitem[Yamada et~al.(2022)Yamada, Kataoka, Chiba, Domae, and
  Ogata]{yamada:2022:fractal_point}
Ryosuke Yamada, Hirokatsu Kataoka, Naoya Chiba, Yukiyasu Domae, and Tetsuya
  Ogata.
\newblock Point cloud pre-training with natural {3D} structures.
\newblock In \emph{Proceedings of the IEEE/CVF Conference on Computer Vision
  and Pattern Recognition}, pages 21283--21293, 2022.

\bibitem[Yu et~al.(2022)Yu, Tang, Rao, Huang, Zhou, and Lu]{yu:2022:point_bert}
Xumin Yu, Lulu Tang, Yongming Rao, Tiejun Huang, Jie Zhou, and Jiwen Lu.
\newblock {Point-BERT}: Pre-training {3D} point cloud transformers with masked
  point modeling.
\newblock In \emph{CVPR}, pages 19313--19322, 2022.

\bibitem[Zhang et~al.(2022{\natexlab{a}})Zhang, Guo, Gao, Fang, Zhao, Wang,
  Qiao, and Li]{zhang:2022:point_m2ae}
Renrui Zhang, Ziyu Guo, Peng Gao, Rongyao Fang, Bin Zhao, Dong Wang, Yu Qiao,
  and Hongsheng Li.
\newblock {Point-M2AE}: Multi-scale masked autoencoders for hierarchical point
  cloud pre-training.
\newblock \emph{arXiv preprint arXiv:2205.14401}, 2022{\natexlab{a}}.

\bibitem[Zhang et~al.(2022{\natexlab{b}})Zhang, Guo, Zhang, Li, Miao, Cui,
  Qiao, Gao, and Li]{zhang:2022:pointclip}
Renrui Zhang, Ziyu Guo, Wei Zhang, Kunchang Li, Xupeng Miao, Bin Cui, Yu Qiao,
  Peng Gao, and Hongsheng Li.
\newblock {PointCLIP}: Point cloud understanding by {CLIP}.
\newblock In \emph{Proceedings of the IEEE/CVF Conference on Computer Vision
  and Pattern Recognition}, pages 8552--8562, 2022{\natexlab{b}}.

\bibitem[Zhang et~al.(2021)Zhang, Girdhar, Joulin, and
  Misra]{zhang:2021:depthcontrast}
Zaiwei Zhang, Rohit Girdhar, Armand Joulin, and Ishan Misra.
\newblock Self-supervised pretraining of {3D} features on any point-cloud.
\newblock In \emph{Proceedings of the IEEE/CVF International Conference on
  Computer Vision}, pages 10252--10263, 2021.

\end{thebibliography}
}

\clearpage
\setcounter{page}{1}
\maketitlesupplementary

\appendix

\section{Object and Scene Numbers}

This work proposes generating scenes using spherical harmonics as objects. The following experiments investigate the impact of object and scene numbers. All experiments are conducted using an MAE. 

The number of used spherical harmonics in scene generation is first fixed (\ie, 10K), and the scene number is scaled. As shown in the upper half of Table~\ref{tab:datagen_numbers}, using more scenes in pre-training is generally beneficial. Also, doubling the scene number leads to a slightly higher AP50~(\SI{0.2}{\percent}) than the default setup. 
However, there is no further improvement when 390K scenes~($\times5$) are applied. 

\begin{table}[b]
    \centering
	\begin{tabular}{llcc}
		\toprule
		\textbf{Objects} & \textbf{Scenes} & \textbf{AP25} (\%) & \textbf{AP50} (\%) \\
		\midrule
		N/A & N/A & 61.6 & 38.8 \\
		\midrule
		$\times$1 & $\times$0.2 & 62.9 & 41.1 \\
		$\times$1 & $\times$0.5 & 63.5 & 42.7 \\
		$\times$1 & $\times$1 & 63.8 & 43.2 \\
		$\times$1 & $\times$2 & 63.8 & \textbf{43.4} \\
		$\times$1 & $\times$5 & 63.4 & 42.9 \\
		\midrule
		$\times$0.2 & $\times$1 & 62.4 & 41.8 \\
		$\times$0.5 & $\times$1 & \textbf{63.9} & 42.8 \\
		$\times$1 & $\times$1 & 63.8 & 43.2 \\
		$\times$2 & $\times$1 & 63.2 & 41.7 \\
		$\times$5 & $\times$1 & 62.5 & 40.6 \\
		\midrule
		$\times2$ & $\times2$ & 63.2 & 42.5 \\
		$\times5$ & $\times5$ & 63.0 & 42.2 \\
		\bottomrule
	\end{tabular}
	\caption{Impact of object and scene numbers. All experiments use multi-view point clouds generated with spherical harmonics. Models are evaluated for object detection on ScanNet. $\times1$: 10K objects or 78K scenes, respectively (\ie, default numbers). $\times a$: scaling with factor $a$. }
	\label{tab:datagen_numbers}
\end{table}

Furthermore, the scene number is fixed to the default value, and the object number is scaled. 
With a small object number ($\times0.2$), the generated scenes have a smaller diversity, which harms the effect of pre-training. However, a large object number (\eg, $\times5$) leads to worse results than the default setup. 
Also, scaling up the scene and object numbers simultaneously~(the last two rows) does not bring improvement. 
It is probably due to the fact that the diversity of spherical harmonics described by Equation~\ref{eq:harmonics} is bounded. 
Since the equation has only eight coefficients, with a too large object number, the minimal ``distance'' between generated objects becomes smaller\footnote{Consider the situation where $n$ points in a continuous interval [0, 1] have to be independently and uniformly sampled. The minimal distance between two points decreases with increasing $n$.}. As a result, some generated scenes contain very similar objects, which hurts the effectiveness of pre-training. 

Based on the discussion, searching for new parametric or formula-driven shapes with more degrees of freedom is beneficial. We leave the exploration to future works.

\section{Scene Generation Rules}
\label{subsec:datagen_rules}
We generate randomized scenes using the rules in \cite{rao:2021:randomrooms}. Some modifications are made to the original configurations. The resulting procedure can be summarized as follows: 

\begin{enumerate}
	\item Randomly pick 12 to 16 objects from the object set. 
	\item Independently apply data augmentation to each object (more details in Section~\ref{subsec:datagen_dataaug}).
	\item Sort the objects in descending order based on their projection areas on the horizontal plane (X-Y plane). 
	\item Randomly set the area (X-Y plane) of a rectangular room depending on the area sum of objects so that the objects can fit into the generated scene. 
	\item Randomly generate walls and floors according to the room area. 
	\item Place each object by randomly choosing a position on the X-Y plane. 
	Objects can be stacked on previous ones as long as the current height of the position is below \SI{2}{\meter}. Otherwise, choose a new position.  Objects cannot overlap.
	\item For multi-view point clouds, voxelize the entire scene with voxel size \SI{0.04}{\meter}. Then, randomly sample 40000 points and save them for pre-training.
	\item For single-view point clouds, randomly place a virtual camera in the scene and capture a depth map via ray-casting. Check the result so that each depth map contains at least 7 objects. All depth maps have the resolution 640$\times$480, following the real-world dataset ScanNet~\cite{dai:2017:scannet_dataset}. The camera intrinsics and camera pose are also saved. 
\end{enumerate}

\section{Pre-Processing and Data Augmentation}
\label{subsec:datagen_dataaug}
The pre-processing and data augmentation in this work are explained in the following. 

\subsection{Data Generation}
Data augmentation is applied to each object during scene generation. Specifically, each object is first normalized into a unit sphere and is randomly scaled with a factor uniformly distributed in $[0.7, 1.5]$. 
Then the object is left-right flipped with a probability of 0.5 and is rotated around the vertical axis (Z-axis) with an arbitrary angle. 

For spherical harmonics, the Z-axis and Y-axis are swapped with the probability 0.5 since spherical harmonics show approximate rotational symmetry around Z-axis~(see Figure~\ref{fig:datagen_harmonics_p1} and \ref{fig:datagen_harmonics_m1}). The motivation is to prevent models from learning the bias that all objects are approximately symmetric around the vertical axis. 
At last, 3000 points are randomly sampled from each object for generating multi-view point clouds. 
For generating single-view data via ray-casting, the mesh representation is directly used without sampling. 

\subsection{Pre-Training}
The primary pre-processing involved in the pre-training pipeline is cropping regions from an entire scene. However, the generated data must be differently pre-processed depending on pre-training methods and inputs: 

\begin{enumerate}
	\item For MAE with multi-view point clouds: A random point is picked from a complete point cloud, and the closest 20000 points (\ie, \SI{50}{\percent}) around it are cropped. 
	\item For MAE with single-view point clouds: A rectangular region with a random ratio in $[0.6, 0.8]$ is cropped from a depth map and transformed into a point cloud. 
	\item For contrastive learning: Similar approaches are applied as for MAE. This work crops spherical regions in multi-view point clouds and crops depth maps for single-view point clouds. However, two overlapping crops are generated in each scene instead of one to create a positive pair for contrastive learning. 
\end{enumerate}

Subsequently, standard data augmentation is applied to each crop, including random translation, rotation around the Z-axis, scaling, point jitter, and left-right flipping.

\section{Generating Spherical Harmonics}

To generate sufficient objects for creating randomized scenes, we randomly set the coefficients in spherical harmonics described in Equation~\ref{eq:harmonics}. 

\begin{figure}[hbt]
	\centering
	\setlength\tabcolsep{2pt}
	\begin{tabular}{cccc}
		\includegraphics[width=0.22\linewidth]{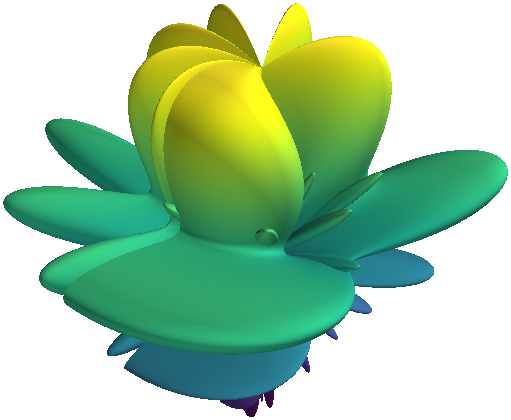} & 
		\includegraphics[width=0.22\linewidth]{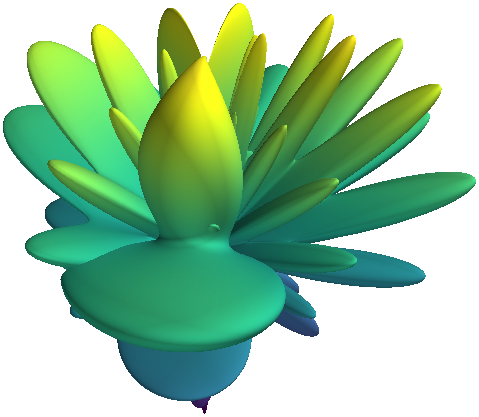} &
		\includegraphics[width=0.22\linewidth]{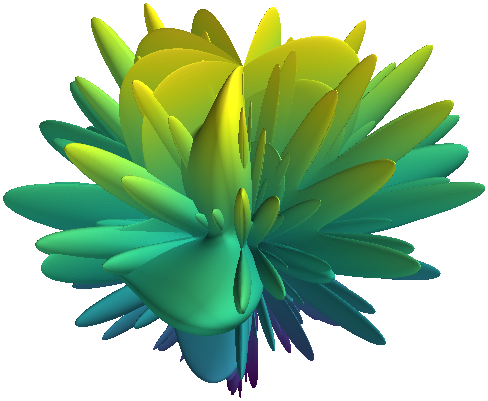} &
		\includegraphics[width=0.22\linewidth]{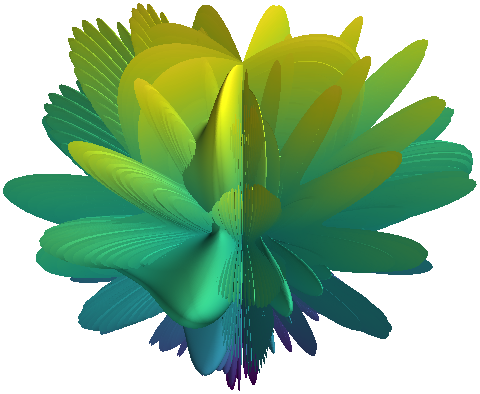} \\
		$p_1 = 0$ & $p_1 = 1$ & $p_1 = 2$ & $p_1 = 3$ \\
		\includegraphics[width=0.22\linewidth]{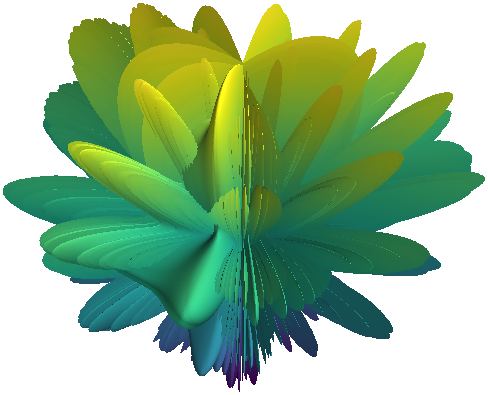} & 
		\includegraphics[width=0.22\linewidth]{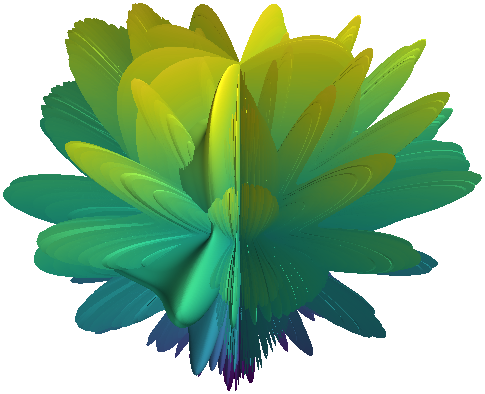} &
		\includegraphics[width=0.22\linewidth]{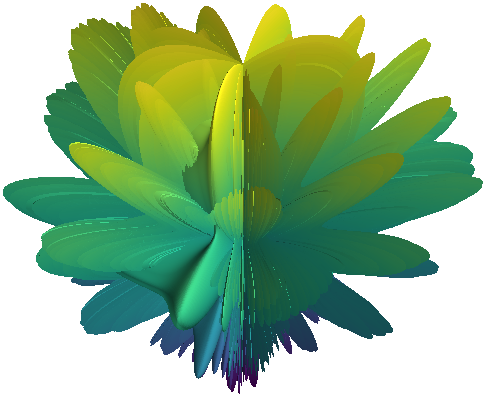} &
		\includegraphics[width=0.22\linewidth]{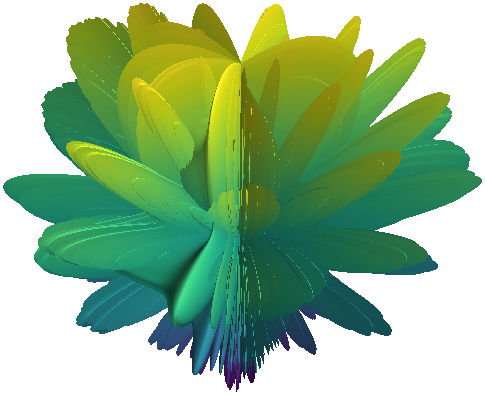} \\
		$p_1 = 4$ & $p_1 = 5$ & $p_1 = 6$ & $p_1 = 7$
	\end{tabular}
	\caption{Impact of coefficient $p_1$ in spherical harmonics. For all objects: $[m_1, m_2, m_3, m_4] = [2, 1, 2, 2]$ and $[p_2, p_3, p_4] = [2, 1, 2]$. }
	\label{fig:datagen_harmonics_p1}
\end{figure}

\begin{figure}[hbt]
	\centering
	\setlength\tabcolsep{2pt}
	\begin{tabular}{cccc}
		\includegraphics[width=0.22\linewidth]{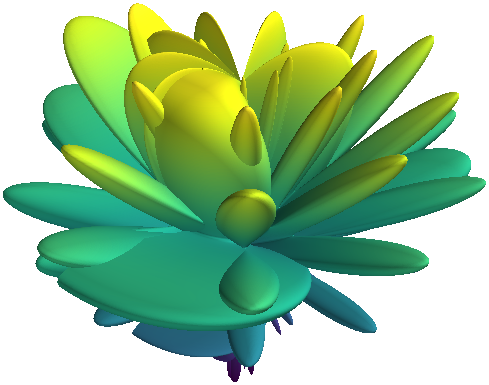} & 
		\includegraphics[width=0.22\linewidth]{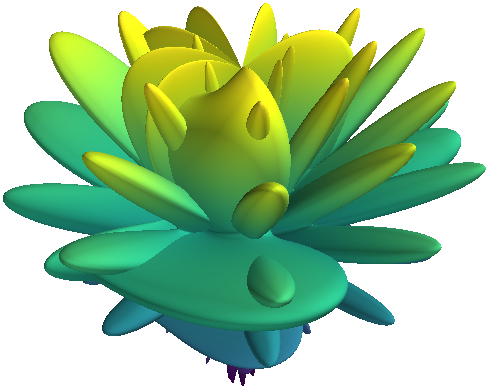} & 
		\includegraphics[width=0.22\linewidth]{figures/coeff/002/2_2_1_2_2_1_2_2.png} & 
		\includegraphics[width=0.22\linewidth]{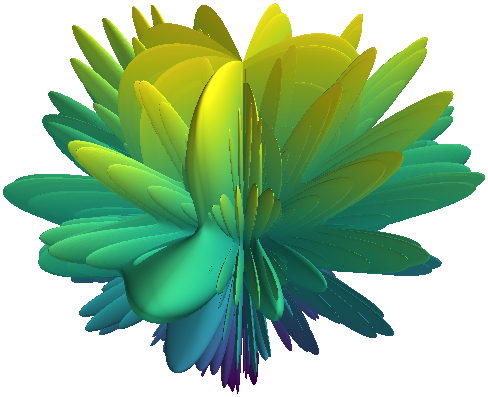} \\
		$m_1 = 0$ & $m_1 = 1$ & $m_1 = 2$ & $m_1 = 3$ \\
		\includegraphics[width=0.22\linewidth]{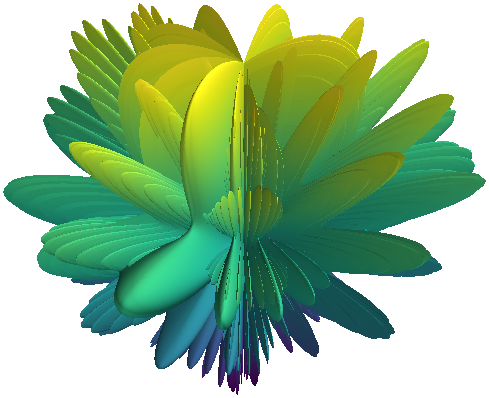} & 
		\includegraphics[width=0.22\linewidth]{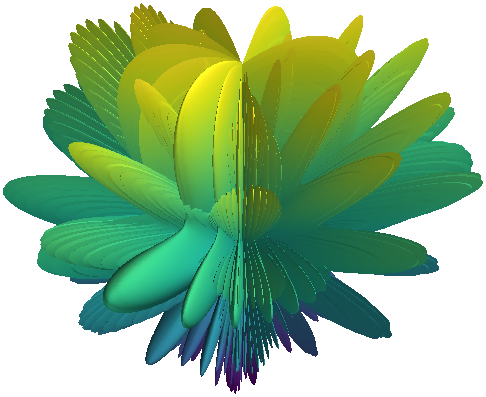} & 
		\includegraphics[width=0.22\linewidth]{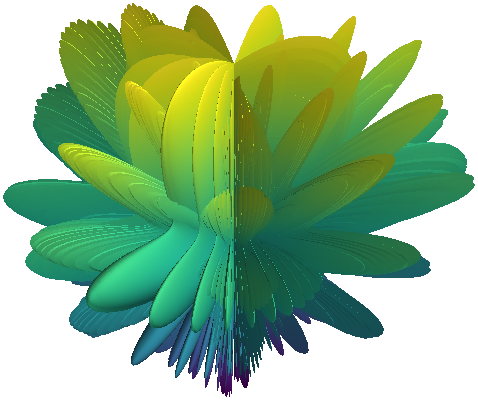} & 
		\includegraphics[width=0.22\linewidth]{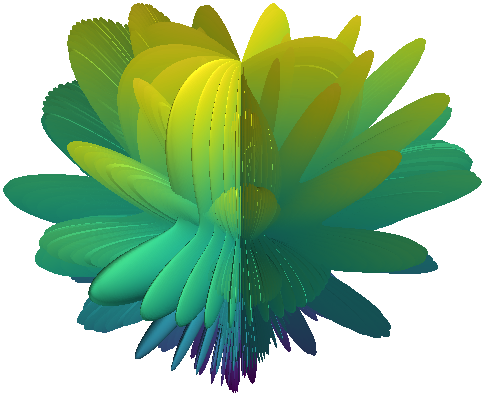} \\
		$m_1 = 4$ & $m_1 = 5$ & $m_1 = 6$ & $m_1 = 7$
	\end{tabular}
	\caption{Impact of coefficient $m_1$ in spherical harmonics. For all objects: $[m_2, m_3, m_4] = [1, 2, 2]$ and $[p_1, p_2, p_3, p_4] = [2, 2, 1, 2]$. }
	\label{fig:datagen_harmonics_m1}
\end{figure}

The coefficients $m_i$ are constrained in the range $[-5,~5]$ and $p_i$ in $[0,~ 4]$, because too large coefficients lead to high-frequency structures.
Due to the sampling theorem, these details are lost when points are sampled from spherical harmonics (represented as meshes) and are thus invisible in the pre-training. 

The impact of the coefficients is illustrated in Figure~\ref{fig:datagen_harmonics_p1}, where all coefficients are fixed except $p_1$. With increasing $p_1$, the ``fins'' of the corresponding spherical harmonics become finer. A similar effect can be seen in Figure~\ref{fig:datagen_harmonics_m1}, where the influence of $m_1$ is visualized. With large coefficients, the overall appearance of spherical harmonics is similar. 
Thus, we constrain the spherical harmonics in the low-frequency range to generate diverse scenes.

\section{More Generated Scenes}
\begin{figure*}
	\centering
	\setlength\tabcolsep{2pt}
	\begin{tabular}{cc}
		\includegraphics[width=0.35\linewidth]{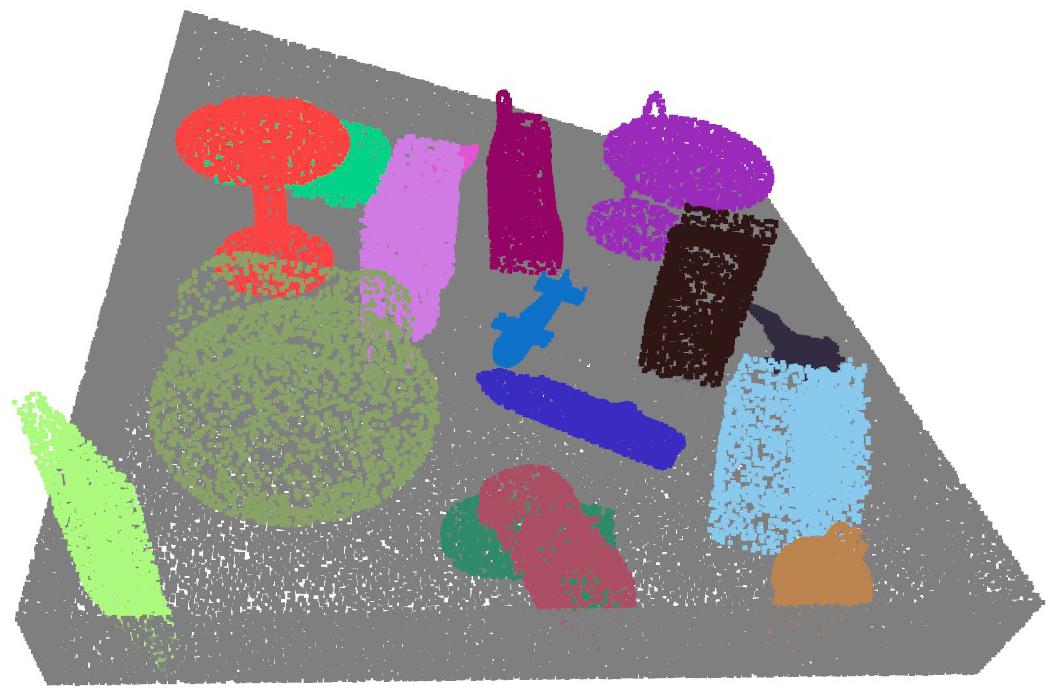} &
		\includegraphics[width=0.25\linewidth]{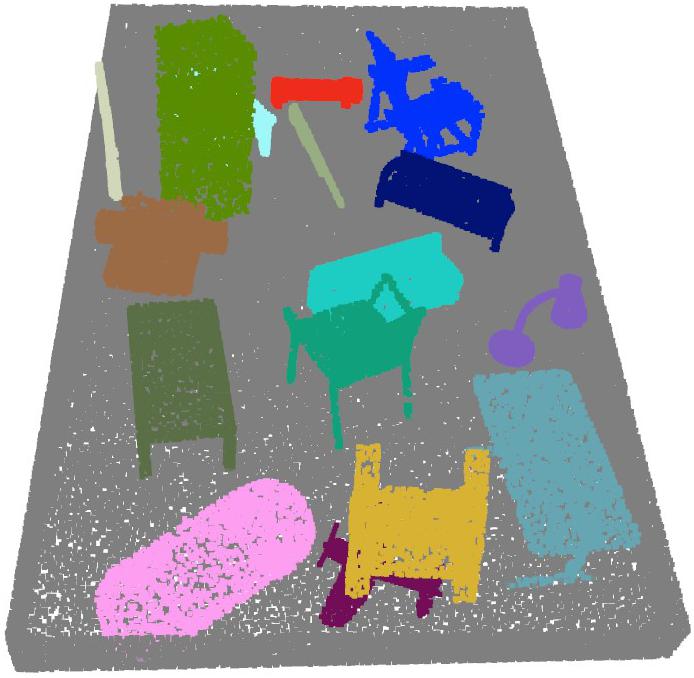} \\
		\includegraphics[width=0.40\linewidth]{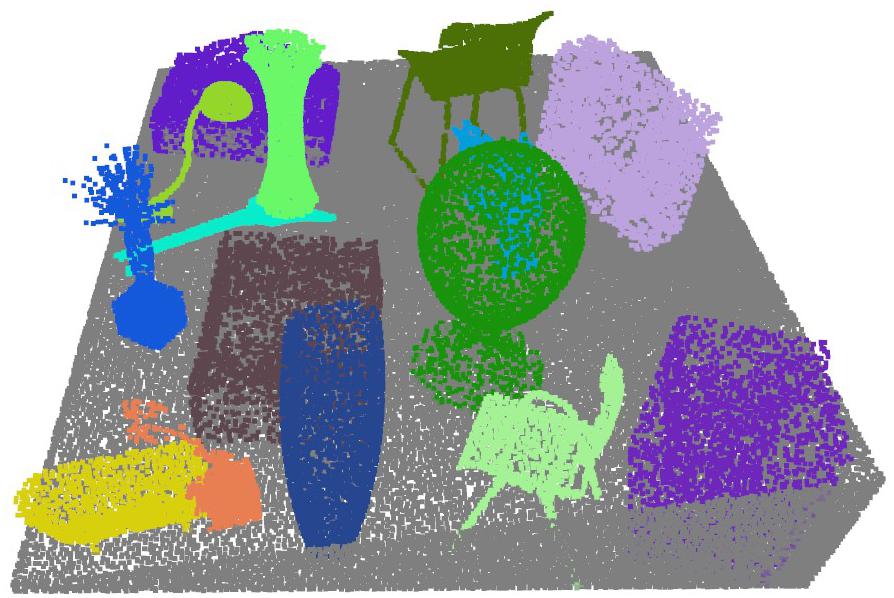} & 
		\includegraphics[width=0.30\linewidth]{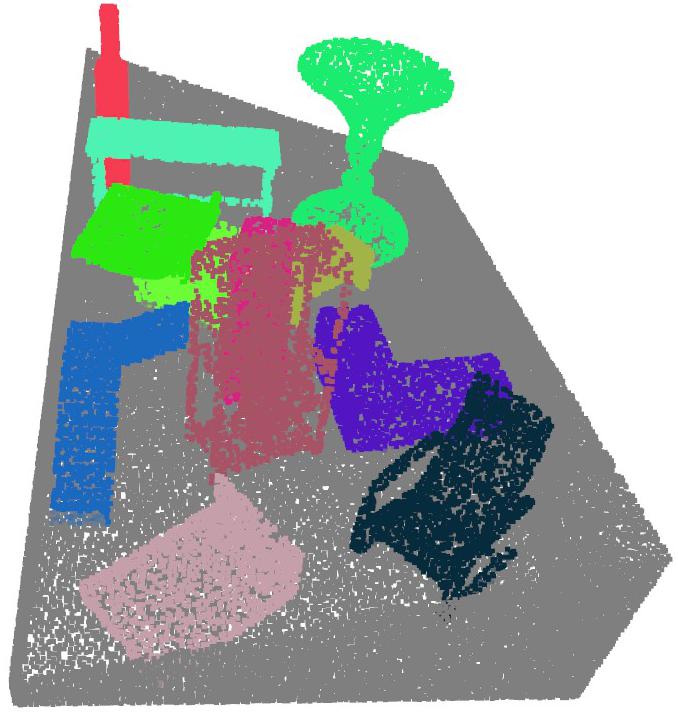} \\
		\includegraphics[width=0.35\linewidth]{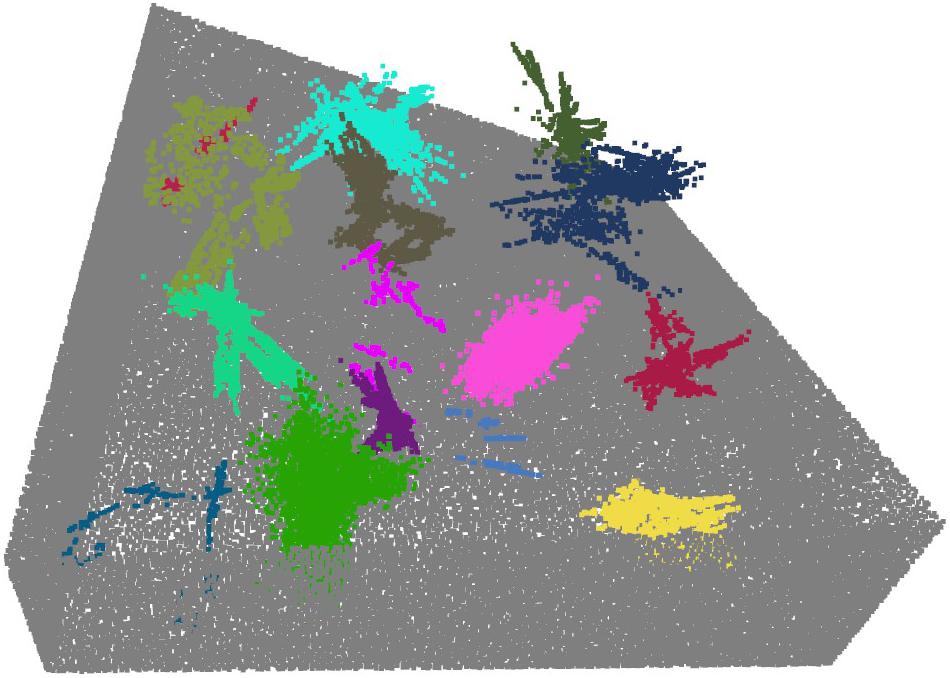} & 
		\includegraphics[width=0.45\linewidth]{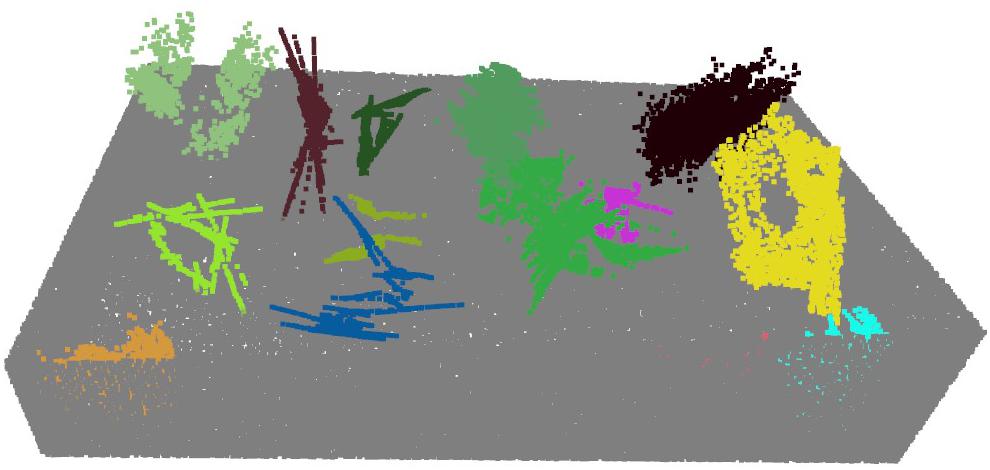} \\
		\includegraphics[width=0.27\linewidth]{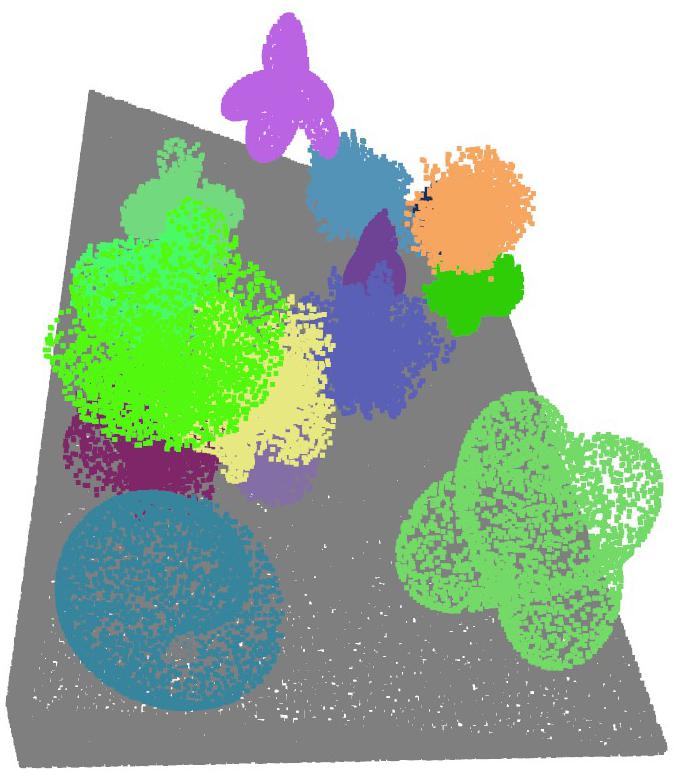} &
		\includegraphics[width=0.45\linewidth]{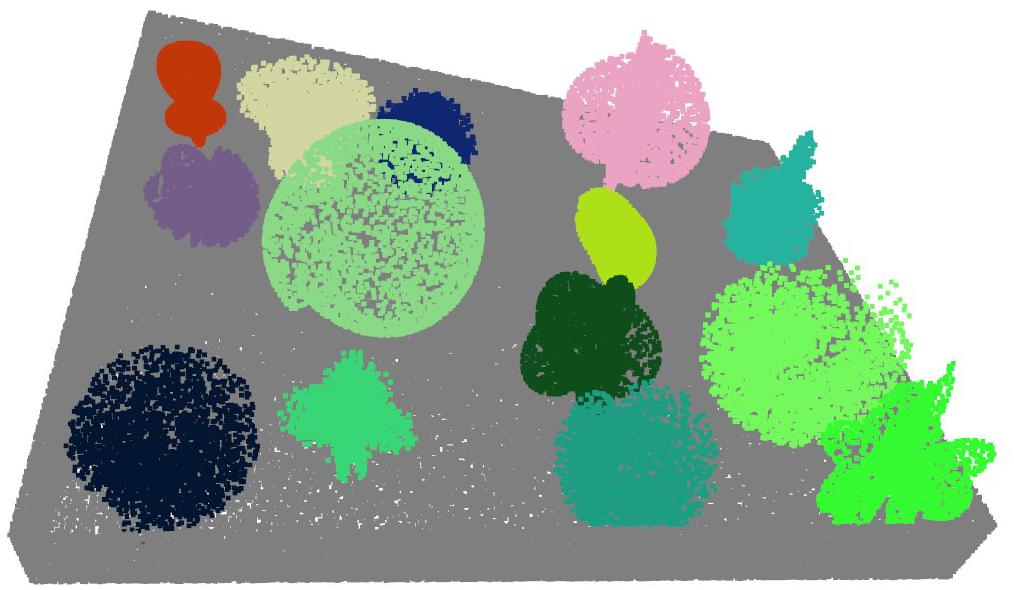} \\
	\end{tabular}
	\caption{More examples of generated scenes. From the first to the fourth row: RM-ShapeNet, RM-ModelNet, RM-Fractal, and RM-Harmonics, respectively. }
	\label{fig:more_scenes}
\end{figure*}

More generated scenes are visualized in Figure~\ref{fig:more_scenes}.

\end{document}